%% file: main.tex
\documentclass[10pt,journal,compsoc]{IEEEtran}
\usepackage{graphicx}

\usepackage{tikz}
\usepackage{comment}
\usepackage{amsmath,amssymb} 
\usepackage{color}
\usepackage{colortbl}
\usepackage{booktabs}
\usepackage{multirow}
\usepackage{wrapfig}
\usepackage{graphicx}
\usepackage{subfigure}
\usepackage{pifont}

\usepackage{url}
\input{math_commands}

\usepackage{xcolor}
\usepackage{orcidlink}
\usepackage{marvosym}

\usepackage{algorithm}
\usepackage{algorithmic}

\def\framework{SceneDreamer}
\def\scenerepresentation{BEV scene representation}
\definecolor{rred}{RGB}{245, 152, 153}
\definecolor{oorange}{RGB}{253, 205, 154}
\definecolor{yyellow}{RGB}{255, 255, 153}
\definecolor{lightorange}{RGB}{251, 229, 214}

\newcommand{\eg}{\textit{e.g.,~}}
\newcommand{\ie}{\textit{i.e.,~}}
\newcommand{\cmark}{\color{green}\ding{51}}%
\newcommand{\xmark}{\color{red}\ding{55}}%
\renewcommand{\paragraph}[1]{\noindent\textbf{#1}}
\ifCLASSOPTIONcompsoc
  \usepackage[nocompress]{cite}
\else
  \usepackage{cite}
\fi

\ifCLASSINFOpdf
\else
\fi
\begin{document}
%
\title{SceneDreamer: Unbounded 3D Scene Generation from 2D Image Collections}

%
%

\author{Zhaoxi Chen,
        Guangcong Wang, 
        and Ziwei Liu
\IEEEcompsocitemizethanks{\IEEEcompsocthanksitem Z. Chen, G. Wang, and Z. Liu are affiliated with S-Lab, Nanyang Technological University.
}
}

%
%

\markboth{ }%
{Shell \MakeLowercase{\textit{et al.}}: Bare Demo of IEEEtran.cls for Computer Society Journals}
%



\input{sections/00_abstract}

\maketitle

\IEEEdisplaynontitleabstractindextext

%
\IEEEpeerreviewmaketitle

\input{sections/01_introduction}
\input{sections/02_review}
\input{sections/03_method}

\input{sections/04_experiment}
\input{sections/05_conclusion}


\ifCLASSOPTIONcompsoc
  \section*{Acknowledgments}
\else
  \section*{Acknowledgment}
\fi

This work is supported by the National Research Foundation, Singapore under its AI Singapore Programme (AISG Award No: AISG2-PhD-2021-08-019), NTU NAP, MOE AcRF Tier 1 (RG13/21), MOE AcRF Tier 2 (T2EP20221-0012), and under the RIE2020 Industry Alignment Fund - Industry Collaboration Projects (IAF-ICP) Funding Initiative, as well as cash and in-kind contribution from the industry partner(s).

\ifCLASSOPTIONcaptionsoff
  \newpage
\fi



%
\bibliographystyle{IEEEtran}
\bibliography{egbib}

%

\begin{IEEEbiography}[{\includegraphics[width=1in,height=1.25in,clip,keepaspectratio]{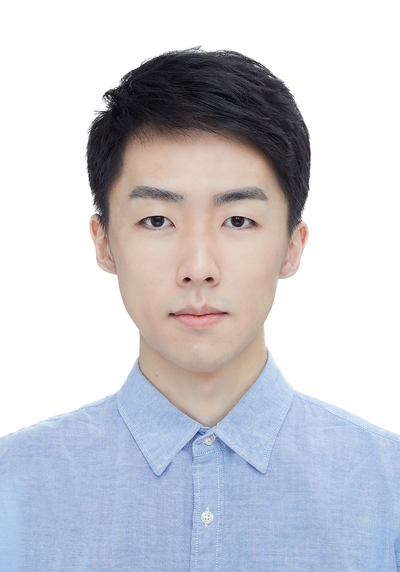}}]{Zhaoxi Chen} is currently a Ph.D. student at MMLab@NTU, Nanyang Technological University, supervised by Prof. Ziwei Liu. He got his bachelor's degree from Tsinghua University in 2021. He received the AISG PhD Fellowship in 2021. His research interests include inverse rendering and 3D generative models. He has published several papers in CVPR, ICCV, ECCV, ICLR and TOG. He also served as a reviewer for CVPR, ICCV, NeurIPS, TOG and IJCV. He is a member of IEEE.
\end{IEEEbiography}

\begin{IEEEbiography}[{\includegraphics[width=1in,height=1.25in,clip,keepaspectratio]{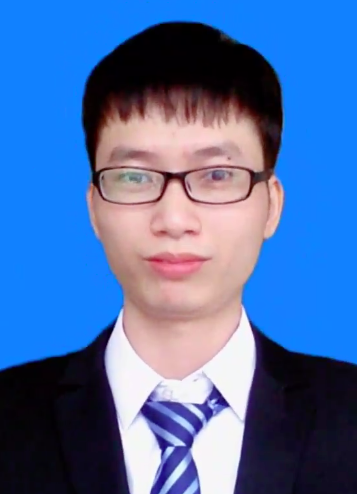}}]{Guangcong Wang} is currently a research fellow in the School of Computer Science and Engineering at Nanyang Technological University, Singapore. He received a Ph.D. degree in the School of Computer Science and Engineering, Sun Yat-sen University, Guangzhou, China, in 2020. His research interests include generative models, 3D, person re-identification, semi-supervised learning, and unsupervised learning. He has published some papers, such as TOG, TNNLS, TIP, TCSVT, ICCV, CVPR, ECCV, KDD, and AAAI. He has served as a reviewer of TPAMI, IJCV, TIP, TCSVT, TOMM, ICML, NeurIPS, ICCV, CVPR, ECCV, ICLR, and AAAI. He is a member of IEEE.
\end{IEEEbiography}

\begin{IEEEbiography}[{\includegraphics[width=1in,height=1.25in,clip,keepaspectratio]{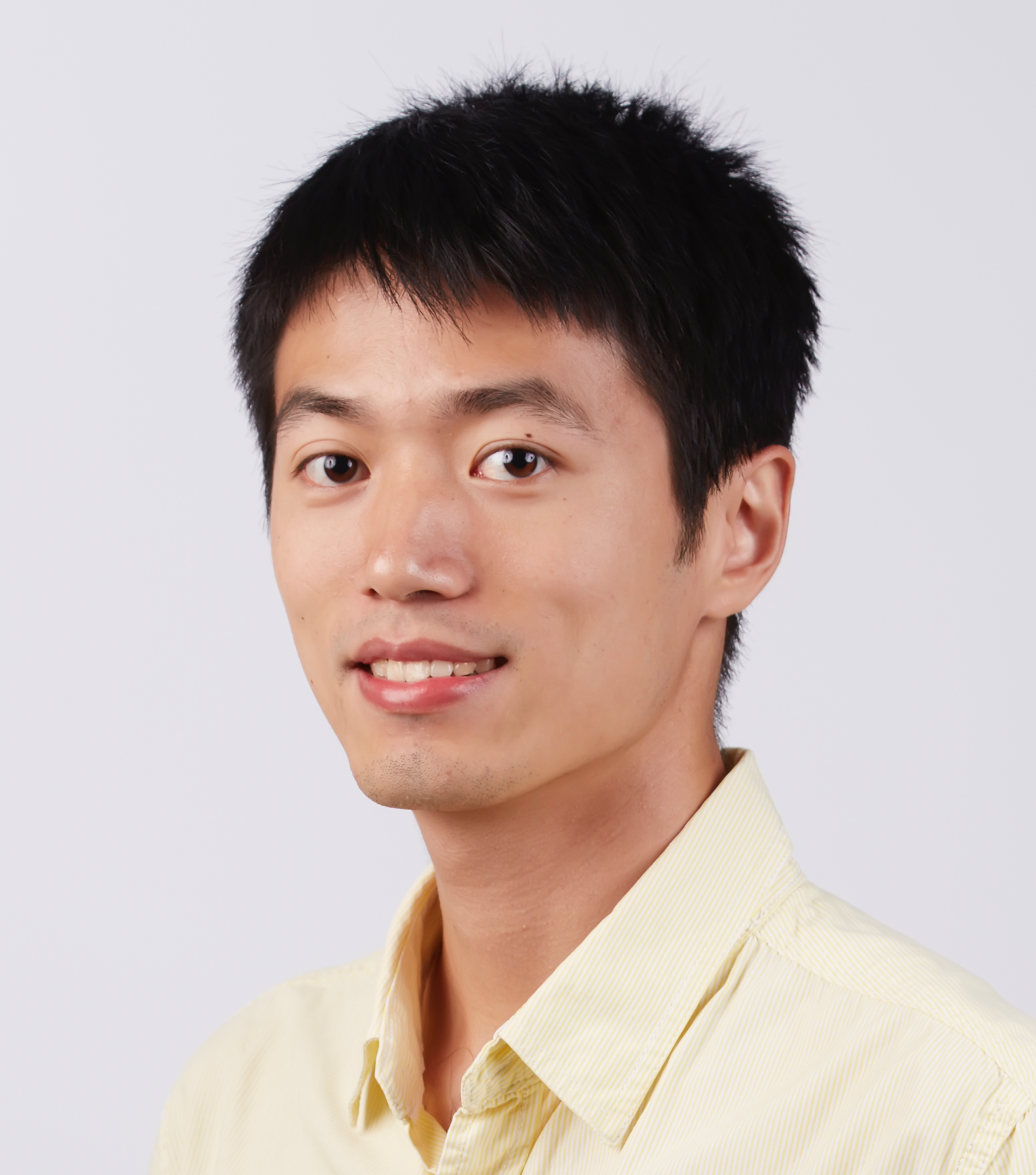}}]{Ziwei Liu}(Member, IEEE) is currently a Nanyang Assistant Professor at Nanyang Technological University, Singapore. His research revolves around computer vision machine learning and computer graphics. He has published extensively on top-tier conferences and journals in relevant fields, including CVPR, ICCV, ECCV, NeurlPS, ICLR, ICML, TPAMI, TOG and Nature - Machine Intelligence. He is the recipient of Microsoft Young Fellowship, Hong Kong PhD Fellowship, ICCV Young Researcher Award, HKSTP Best Paper
Award and WAIC Yunfan Award. He serves as an Area Chair of CVPR, ICCV, NeurlPS and ICLR, as well as an Associate Editor of IJCV.
\end{IEEEbiography}








\end{document}

%% file: math_commands.tex
\usepackage{amsmath,amsfonts,bm}

\def\eqref#1{equation~\ref{#1}}

\def\1{\bm{1}}

\DeclareMathAlphabet{\mathsfit}{\encodingdefault}{\sfdefault}{m}{sl}
\SetMathAlphabet{\mathsfit}{bold}{\encodingdefault}{\sfdefault}{bx}{n}



%% file: sections/00_abstract.tex
\IEEEtitleabstractindextext{%
\begin{abstract}
In this work, we present \textbf{\framework}, an unconditional generative model for unbounded 3D scenes, which synthesizes large-scale 3D landscapes from random noise. Our framework is learned from in-the-wild 2D image collections only, without any 3D annotations. 
At the core of \framework\ is a principled learning paradigm comprising \textbf{1)} an efficient yet expressive 3D scene representation, \textbf{2)} a generative scene parameterization, and \textbf{3)} an effective renderer that can leverage the knowledge from 2D images.
Our approach begins with an efficient bird's-eye-view (BEV) representation generated from simplex noise, which includes a height field for surface elevation and a semantic field for detailed scene semantics. This BEV scene representation enables 1) representing a 3D scene with quadratic complexity, 2) disentangled geometry and semantics, and 3) efficient training. Moreover, we propose a novel generative neural hash grid to parameterize the latent space based on 3D positions and scene semantics, aiming to encode generalizable features across various scenes. Lastly, a neural volumetric renderer, learned from 2D image collections through adversarial training, is employed to produce photorealistic images. Extensive experiments demonstrate the effectiveness of \framework\ and superiority over state-of-the-art methods in generating vivid yet diverse unbounded 3D worlds. Project Page is available at \url{https://scene-dreamer.github.io/}. Code is available at \url{https://github.com/FrozenBurning/SceneDreamer}.

\end{abstract}

\begin{IEEEkeywords}
Neural Rendering, GAN, 3D Generative Model, Unbounded Scene Generation
\end{IEEEkeywords}}

%% file: sections/01_introduction.tex
\begin{figure*}
    \centering
    \includegraphics[width=\textwidth]{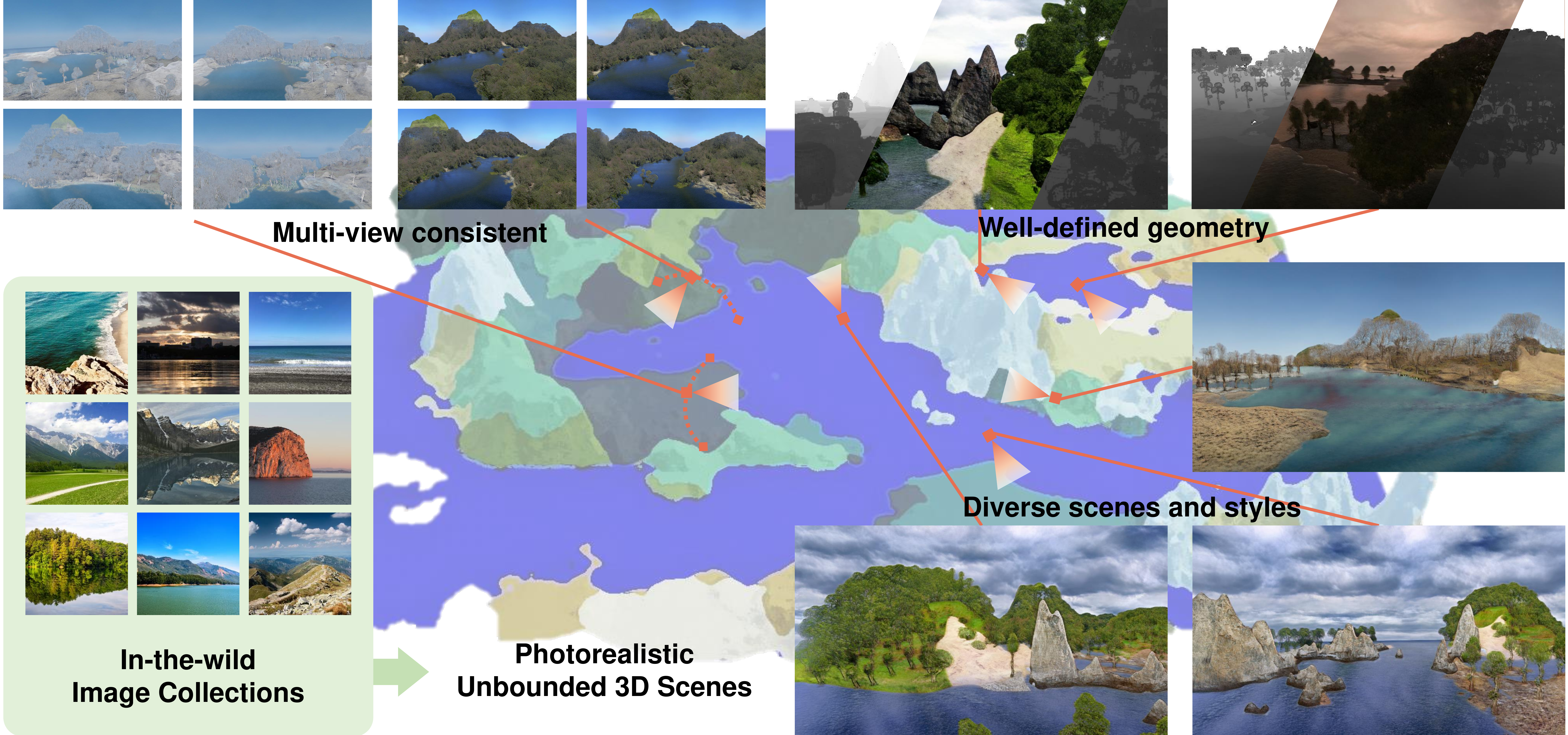}
    \caption{\textbf{\framework\ learns to generate unbounded 3D scenes from in-the-wild 2D image collections.} Our method can synthesize diverse landscapes across different styles, with 3D consistency, well-defined depth, and free camera trajectory.}\label{fig:teaser}
    \vspace{-0.1in}
\end{figure*}

\IEEEraisesectionheading{\section{Introduction}\label{sec:intro}}

%
%
%
%
\IEEEPARstart{S}{cene Generation}
has raised considerable attention in recent years, addressing the growing need for 3D creative tools in the metaverse. At the core of 3D content creation is inverse graphics, which aims to recover 3D representations from 2D observations. Given the cost and labor for creating 3D assets, the ultimate goal of 3D content creation would be learning a generative model from in-the-wild 2D images. Recent work on 3D-aware generative models tackles the problem to some extent, by learning to generate object-level content (\eg faces~\cite{chan_efficient_2022, gu2021stylenerf, or-el_stylesdf_2021}, bodies~\cite{EVA3D} or common objects~\cite{xue_giraffe_2022, gao2022get3d}) from curated 2D image data. However, the observation space is in a bounded domain and the generated target occupies a limited region of Euclidean space. It is highly desirable to learn 3D generative models for unbounded scenes from in-the-wild 2D images, \eg a vivid landscape that covers an arbitrarily large region (as shown in Fig.~\ref{fig:teaser}), which we aim to tackle in this paper.

To generate unbounded 3D scenes from in-the-wild 2D images, intuitively, three critical issues must be addressed: the unbounded range of scenes, unaligned content in scale and coordinate, and in-the-wild 2D images without knowledge of camera pose. Specifically, a successful unbounded scene generation model should overcome the following challenges: \textbf{1)} Lack of efficient 3D representation for unbounded 3D scenes. Unbounded scenes can occupy an arbitrarily large region of Euclidean space, necessitating the design of an efficient 3D representation; \textbf{2)} Lack of content alignment. Given a set of in-the-wild 2D images, objects with different semantics could be captured in varying scales, 3D locations, and orientations. The unaligned content often leads to unstable training; \textbf{3)} Lack of priors on camera pose distributions. In-the-wild 2D images may originate from non-overlapping views or different image sources, making it difficult to use structure-from-motion to estimate the camera poses on in-the-wild 2D images due to the absence of reliable correspondence between different images.

Existing methods only offer partial solutions to these three challenges, as summarized in Table~\ref{tab:motivation}. For the first problem, object-centric 3D GANs~\cite{chan_efficient_2022, chan_pi-gan_2021, gu2021stylenerf} warp the scene to a bounded domain within the camera view. However, the space and scale of occupancy are not determined for unbounded scenes. Plus, instead of capturing the entire scene, observation space in unbounded scenes is often a partial and small portion, which makes it a waste of capacity to model the scene as a whole. Though efficient representations, \eg tri-plane~\cite{chan_efficient_2022}, 2D manifolds~\cite{deng_gram_nodate, xiang_gram-hd_2022}, sparse voxel grid~\cite{schwarz_voxgraf_2022} have been proposed for object-centric 3D GANs, they model the scene as a whole which is not practical for large-scale unbounded scenes. For the second problem, existing 3D GANs either need ground truth camera poses~\cite{devries_unconstrained_2021} or require a camera pose prior~\cite{chan_efficient_2022}. However, we cannot estimate camera pose distribution on in-the-wild images collected from non-overlapping views or different image sources. For the third problem, current 3D GANs are trained on curated and aligned datasets~\cite{liuLQWTcvpr16DeepFashion, karras_style-based_2019, shapenet2015} where all objects are in the same scale, 3D location, and orientation and the camera covers the entire scene. However, in-the-wild scene-level content may lead to unstable training due to its extremely high diversity. It is crucial for 3D generative models to align in-the-wild 2D content with 3D semantics. 

Given the aforementioned challenges, we propose a principled learning paradigm, \framework, that learns to generate unbounded 3D scenes from in-the-wild 2D image collection without camera parameters. To facilitate that, our framework consists of three modules, an efficient yet expressive 3D scene representation, a generative scene parameterization, and a volumetric renderer that can leverage knowledge from 2D images.

First, we propose a \scenerepresentation\ that is both efficient and expressive for 3D unbounded scenes (Challenge 1). Given a simplex noise as input, we generate a height field and a semantic field to represent a unique 3D scene from a bird's-eye-view. The height field represents the surface elevation of the scene, while the semantic field provides detailed scene semantics. The advantages of such a representation are three folds. \textbf{1)} It enables learning 3D from in-the-wild images without any knowledge of camera poses by explicitly sampling camera views in constructed 3D worlds. \textbf{2)} It supports a sufficient level of detail both in geometry and semantics. \textbf{3)} It represents a 3D scene with quadratic space complexity, which enables efficient training and 3D points sampling. 

Furthermore, we introduce a novel semantic-aware neural hash grid for adversarial training to parameterize the space-varied and scene-varied latent features. Different from the hash grid as neural graphics primitives~\cite{muller_instant_nodate}, we propose a semantic-conditional hash grid that extends the vanilla version into a generative setting by conditioning on \scenerepresentation. It enables our model to learn generalizable latent features across scenes while preserving impressive capacity and efficiency. The proposed generative hash grid is semantic-aware, which further helps the model to align content with 3D semantics (Challenge 2).

Lastly, we employ a style-based volumetric renderer~\cite{hao_gancraft_2021} to produce photorealistic images via adversarial training. Given latent features sampled from the hash grid, the renderer learns to blend them into realistic images by style-modulated volume rendering~\cite{mildenhall_nerf_2020, gu2021stylenerf}. It guarantees the 3D consistency of the generated images while also enabling learning from in-the-wild 2D image collections (Challenge 3). Once training is done, we can unconditionally generate diverse unbounded 3D scenes by sampling different simplex noise and style codes. 

Quantitative and qualitative experiments are performed against various state-of-the-art 3D generative models, demonstrating the capability of \framework\ in generating large-scale and diverse 3D scenes. In conclusion, we summarize our contributions as follows:
\textbf{1)} To the best of our knowledge, we are the first to achieve unbounded 3D scene generation from in-the-wild 2D image collections;
\textbf{2)} We propose a \scenerepresentation\ for an expressive scene modeling and efficient 3D GAN training;
\textbf{3)} We extend the neural hash grid parameterization into the generative setting for the first time;
\textbf{4)} We demonstrate useful applications of \framework, including unconditional 3D landscape generation, perpetual view generation~\cite{liu_infinite_2021}, and interpolation between different scenes and styles.

\begin{table*}[t]
\caption{\textbf{Comparisons of Existing Methods on 3D Scene Generation.} Our aim is to learn a feed-forward generative model of unbounded 3D scenes from in-the-wild 2D image collections. ``Training Data'' denotes the source of training images; ``Rendering Resolution'' denotes the maximum resolution of rendered images during inference. ``Camera DOF'' denotes the degree of freedom of camera movement at inference time, where ``XYZ'' is for translation (x, y, z) and ``RPY'' is for rotation (row, pitch, yaw). ``Unbounded'' means whether the scene can be generated in arbitrary scales. ``3D consistent'' indicates whether the renderings have 3D consistency. ``Feed-Forward'' denotes that the method does not need test-time optimization to generate novel scenes. $^{\dag}$Posed 3D data with image and camera pose pairs.
}\label{tab:motivation}
\centering
\begin{tabular}{cccccccc}
\toprule
     Methods& Training Data & Rendering Resolution & Camera DOF & Unbounded? & 3D Consistent? & Feed-Forward?\\
     \midrule
     GANcraft~\cite{hao_gancraft_2021}& in-the-wild 2D images & $1920\times1080$ & XYZRPY & \xmark & \cmark & \xmark\\
     EG3D~\cite{chan_efficient_2022}&curated 2D images&$512\times512$&RPY&\xmark&\cmark&\cmark\\
     GSN~\cite{devries_unconstrained_2021}&$^{\dag}$3D trajectories&$256\times256$&XYZRPY&\xmark&\cmark&\cmark\\
     Inf-Nature~\cite{liu_infinite_2021}&$^{\dag}$3D trajectories&$160\times256$&XYZ&\cmark&\xmark&\cmark\\
     Inf-Zero~\cite{li_infinitenature-zero_2022}&in-the-wild 2D images&$512\times512$&XYZ&\cmark&\xmark&\cmark\\
     \textbf{Ours}&in-the-wild 2D images&$3840\times2160$&XYZRPY&\cmark&\cmark&\cmark\\
     \bottomrule
\end{tabular}
\vspace{-0.1in}
\end{table*}

%% file: sections/02_review.tex
\section{Related Work}
\label{sec:review}

\paragraph{Neural scene representation.}
Immense progress has been witnessed in neural scene representations~\cite{yu_pixelnerf_nodate, tancik_block-nerf_2022, niemeyer_regnerf_2021, reiser_kilonerf_2021, pumarola_d-nerf_2020, mildenhall_nerf_2021, mildenhall_nerf_2020, park_hypernerf_2021, johari_geonerf_2021, barron_mip-nerf_2021, niemeyer_differentiable_2020, chen_tensorf_2022, muller_instant_nodate, yu_plenoxels_2021, wang_neus_2021, barron_mip-nerf_2021-1, li_mine_nodate, sun_direct_2021, liu_neural_2021} , which can be optimized from 2D multi-view images via differentiable neural rendering~\cite{mildenhall_nerf_2020}. Implicit representations~\cite{mildenhall_nerf_2020, barron_mip-nerf_2021}, typically coordinate networks, employ large MLPs to model the 3D scenes. It offers potential advantages in memory efficiency. However, densely querying the network during training is slow, especially for large-scale unbounded scenes. Explicit representations~\cite{yu_plenoctrees_2021, yu_plenoxels_2021, sun_direct_2021}, \ie voxels, are inherently fast to query. Yet, they are memory-intensive, which makes them difficult to scale up and contain sufficient level of detail in large-scale scenes. Hybrid representations~\cite{chan_efficient_2022, devries_unconstrained_2021, liu_neural_2021, martel_acorn_2021, muller_instant_nodate} combine the benefits of both explicit and implicit representations to strike a balance between efficiency and cost. In terms of balancing the trade-off, the proposed \scenerepresentation\ shares a common goal with them, but ours is unique in its careful design. We explicitly model the geometry and semantics of unbounded scenes within this representation, which is key to efficient training and semantic alignment of content for in-the-wild scene generation.

\paragraph{3D-aware GANs.}
Generative adversarial network (GAN)\cite{NIPS2014_gan} has been a great success in recent years, especially in 2D image generation~\cite{karras_style-based_2019, karras_analyzing_2020}. Extending the capability of GANs to 3D space has emerged as well. A bunch of work~\cite{HoloGAN2019, gadelha3dshape, wu3dgan} intuitively extend the CNN backbone used in 2D to 3D with a voxel-based representation. However, the prohibitively high computational and memory cost of voxel grids and 3D convolution makes them difficult to model unbounded 3D scenes. With recent advances in neural radiance field (NeRF)~\cite{mildenhall_nerf_2020}, many~\cite{niemeyer_giraffe_2021, schwarz_voxgraf_2022, schwarz_graf_2021, skorokhodov_epigraf_nodate, chan_pi-gan_2021, sun_controllable_2022, deng_gram_nodate, xiang_gram-hd_2022, chan_efficient_2022, or-el_stylesdf_2021, gu2021stylenerf, EVA3D, sun2022ide} have incorporated volume rendering as the key inductive bias to make GANs be 3D-aware, which enables GANs to learn 3D representations from 2D images. However, most of them are trained on curated datasets for bounded scenes, \eg human faces~\cite{karras_style-based_2019}, human bodies~\cite{liuLQWTcvpr16DeepFashion}, and objects~\cite{shapenet2015}, where the occupancy of the target in Euclidean space is limited and the camera view captures a complete observation of the scene instead of a partial one. The problem of unbounded 3D scene generation from 2D images is still underexplored. 
As a concurrent work, Persistent Nature~\cite{chai2023persistentnature} leverages persistent layout grids to generate unbounded 3D worlds. Yet, the inefficiency of latent layout grids limits the rendering resolution to $256\times 256$. Meanwhile, InfiniCity~\cite{lin2023infinicity} learns to generate 3D urban scenes with the reliance on 3D CAD datasets.

\paragraph{Scene-level image synthesis.}
Different from impressive 2D generative models which focus on a single category or common objects, generating scene-level content is a challenging task, given the extremely high diversity of scenes. Semantic image synthesis~\cite{park_semantic_2019, esser_taming_2021, mallya_world-consistent_2020, hao_gancraft_2021} is the most promising way for generating scene-level content in the wild. The model is conditioned on pixel-wise dense correspondence, \eg semantic segmentation maps or depth maps, effectively helping the model to find the bijective mapping between input condition and output textures. It is worth mentioned that some~\cite{mallya_world-consistent_2020, hao_gancraft_2021, shi_3d-aware_2022, liu_infinite_2021, li_infinitenature-zero_2022, ren_look_2022} have even succeeded in 3D-aware scene synthesis as well. However, they are neither fully 3D consistent nor support feed-forward generation on novel worlds. Another bunch of work~\cite{paschalidou_atiss_2021, wang_sceneformer_2021, devries_unconstrained_2021, gaudi} focuses on indoor scene synthesis using expensive 3D datasets~\cite{dai2017scannet, replica19arxiv} or CAD retrieval~\cite{fu20213d}, which strictly limits the diversity of their results.

%% file: sections/03_method.tex
\begin{figure*}[t]
    \begin{center}
    \centerline{\includegraphics[width=2.09\columnwidth]{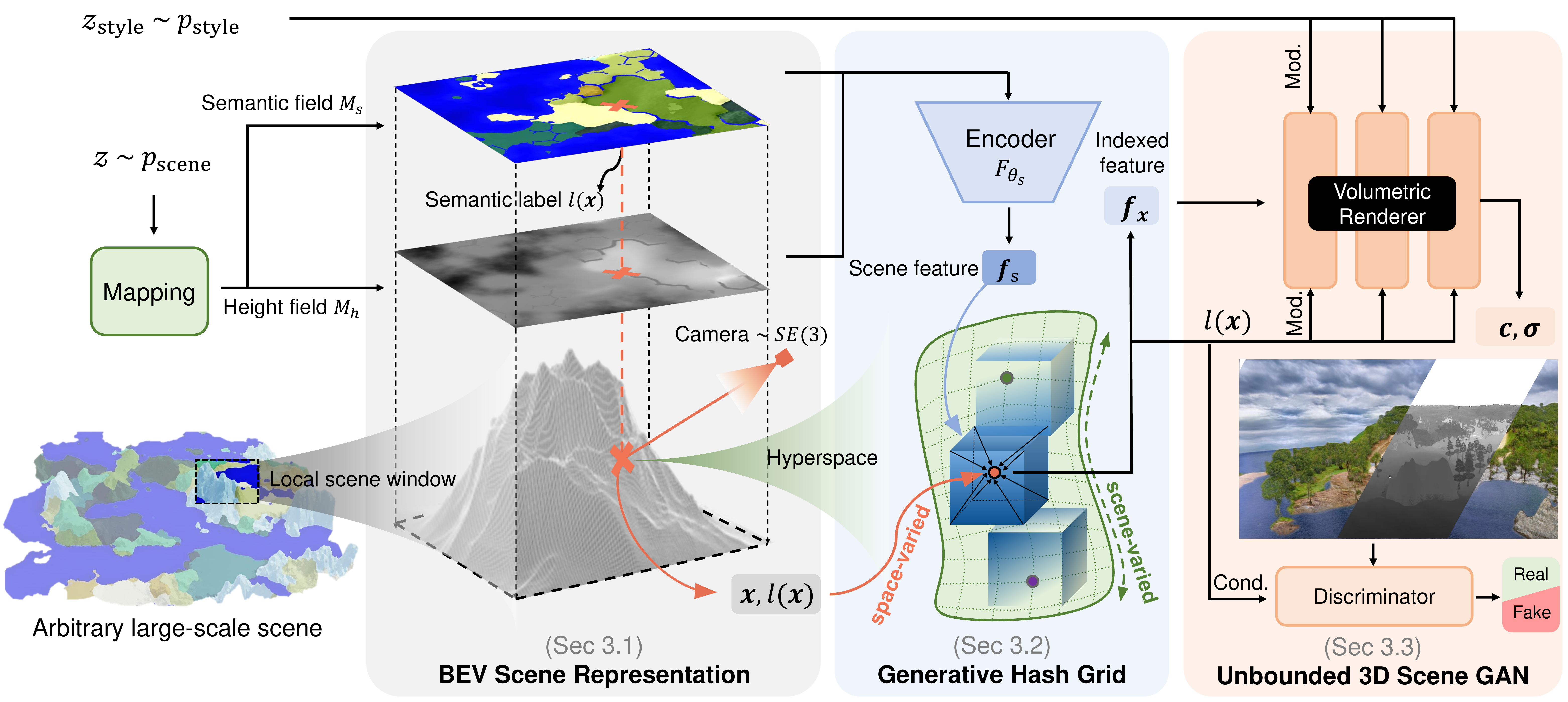}}
    \caption{\textbf{Overview of \framework}. Given a simplex noise $z \sim p_{\mathrm{scene}}$ and a style code $z_\mathrm{style} \sim p_{\mathrm{style}}$ as input, our model is capable of synthesizing large-scale 3D scenes where the camera can move freely and get realistic renderings. We first derive our \scenerepresentation\ which consists of a height field and a semantic field. Then, we use a generative neural hash grid to parameterize the hyperspace of space-varied and scene-varied latent features given scene semantics $\bm{f}_{s}$ and 3D position $\bm{x}$. Finally, a style-modulated renderer is employed to blend latent features $\bm{f}_{\bm{x}}$ and render 2D images via volume rendering.
    }\label{fig:overview}
    \end{center}
    \vspace{-0.1in}
\end{figure*}

\section{SceneDreamer}
\label{sec:method}
Our aim is to learn a feed-forward generative model for unbounded 3D scenes from in-the-wild 2D image collections only. 
We first derive our \scenerepresentation\ (Sec.~\ref{sec:orthogonal}), which consists of a height field and a semantic field, for the aim of efficient 3D point sampling and training. 
Then, we use a generative neural hash grid to parameterize the hyperspace of space-varied and scene-varied latent features (Sec.~\ref{sec:hashgen}), learning generalizable 3D features across scenes.
Finally, a style-based generator is employed to blend latent features of 3D points and render 2D images through volume rendering (Sec.~\ref{sec:nerfgan}). 
An overview is shown in Fig.~\ref{fig:overview}.

\subsection{BEV Scene Representation}
\label{sec:orthogonal}
The learning of unbounded 3D scene GAN requires the underlying 3D representation to be: \textbf{1)} efficient, \ie we can represent a 3D world with relatively low complexity to enable tractable training; \textbf{2)} expressive, \ie we can sample 3D points with corresponding semantics from the representation for ease of semantic content alignment. Existing representations~\cite{deng_gram_nodate, chan_efficient_2022, gu2021stylenerf, schwarz_voxgraf_2022} are designed for object-centric 3D GANs in bounded scenes, and cannot well fit the large scale and diverse content for unbounded scenes. To this end, we propose a novel bird's-eye-view (BEV) scene representation consisting of a height field and a semantic field for large-scale scenes.

\paragraph{Height Field.}
The height field $M_h:\bm{x} \xrightarrow{} h(\bm{x})$ represents the level of height of surface points in a 3D scene ($\bm{x} \in \mathbb{R}^3$). Intuitively, it is similar to the elevation map used in geography, which records the distance of surfaces above zero level. Given a level of detail $D$, \ie the resolution of a tiny voxel, we can further discretize the height field into a height map $\hat{M}_h$ with a resolution of $N^D\times N^D\times 1$. Note that, the spatial resolution $N^D$ can be any positive integer for unbounded scenes and can vary given the budget of resources.

\paragraph{Semantic Field.}
The semantic field $M_s: \bm{x} \xrightarrow{} l_i(\bm{x})$ represents the semantic label of surfaces in a 3D scene, where $i \in \{1, \dots, C_s\}$ and $C_s$ is the number of semantic classes (\eg rivers, mountains, grasslands). Similar to the height map, we can derive a semantic map $\hat{M}_s$ from the semantic field via discretization, with a resolution of $N^D\times N^D\times C_s$. The semantic map stores the one-hot encoding of semantic labels for corresponding surface points in the height map.

With such a representation, we can easily increase memory efficiency during training while preserving a satisfied level of detail by constructing local 3D volumes. Given a query of any 3D position $\bm{x} \in \mathbb{R}^3$, we can first crop a local scene window with a resolution of $N_w^D\times N_w^D$ from the $D$-th level of $\hat{M}_h$ and $\hat{M}_s$. Then, the local 3D voxels $V_w$ with a resolution of $N^D_w \times N^D_w \times H^D_w$ are constructed using the 3D coordinates from height map and semantic labels from the semantic map. The trade-off between memory and capacity during volume rendering (Sec.~\ref{sec:nerfgan}) can be achieved by tuning $N^D$ and $N^D_w$. 

In addition, such a representation facilitates a feed-forward design for large-scale scene generation, where the input latent code $z \sim p_{\mathrm{scene}}$ can be transformed to the scene representation through a mapping: $z \xrightarrow{} (M_h, M_s)$. This mapping can be either parameterized by networks or parameter-free. As for the former way, one can achieve this by learning from satellite images or real terrain data, which we leave for future work. Our instantiation of this mapping is a parameter-free way, which enables us to scale up the training data for free. Inspired by the fact that the simplex noise can mimic patterns of nature~\cite{perlin1985image}, we let $z$ be a randomized simplex noise as the input, and generate the height field and the semantic field accordingly. Please refer to Sec.~\ref{sec:detail} for implementation details.

\begin{figure*}[t]
    \begin{center}
    \centerline{\includegraphics[width=2.0\columnwidth]{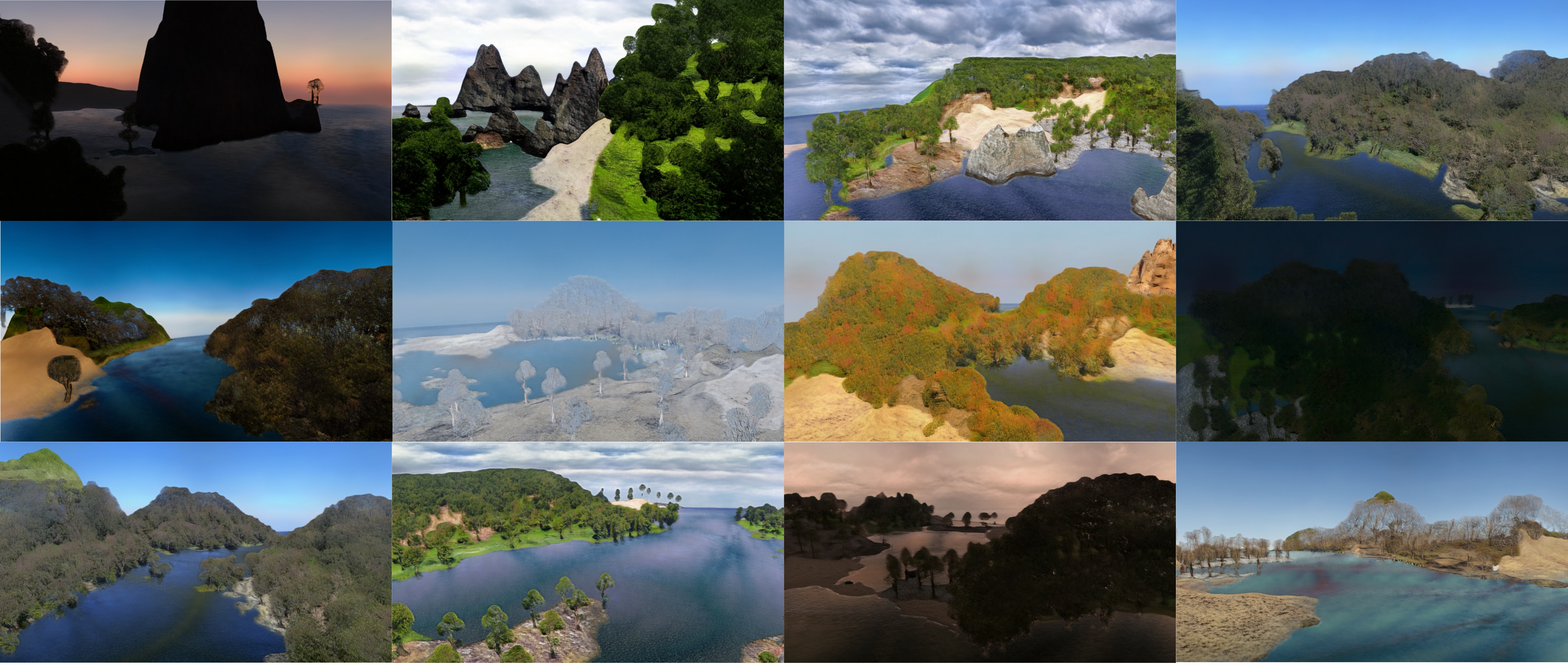}}
    \caption{\textbf{Diverse samples of \framework}. Our model can synthesize a large variety of 3D scenes with diverse styles, from winter to summer and dawn to dusk. Please check the supplementary and project page for 3D consistent videos.}\label{fig:more-sample}
    \end{center}
    \vspace{-0.1in}
\end{figure*}

\subsection{Generative Neural Hash Grid}
\label{sec:hashgen}
To enable generalizable 3D representation learning across scenes and align content with 3D semantics, we need to parameterize the scene representation into latent space for ease of adversarial learning. Given the large scale of unbounded scenes and the fact that only the surface points matter for appearance, it is a waste of capacity to model the entire 3D space, \eg using 3D convolutions~\cite{HoloGAN2019} or tri-plane~\cite{chan_efficient_2022}. Plus, the observation space involved in a camera view is often a small portion of the unbounded scene, which suggests that an ideal parameterization should be able to activate the related subspace instead of the whole scene. Inspired by the success of neural hash grid~\cite{muller_instant_nodate} in single-scene reconstruction, we take advantage of this flexible yet compact parameterization and extend it to learn generalizable features across scenes by modeling the hyperspace beyond 3D space. 

The vanilla neural hash grid represents a hash-based encoding function $\mathrm{enc}(\bm{x}; \theta_H): \mathbb{R}^3 \xrightarrow{} \mathbb{R}^{C_H}$, which maps the input $\bm{x}$ to a unique indexed feature with $C_H$ dimensions in the trainable encoding parameters $\theta_H$. We extend it to a generative setting by introducing a hyperspace, aiming to learn space-varied and scene-varied features. In specific, we first encode a sampled scene $(\hat{M}_h, \hat{M}_s)$, using an encoder $F_{\theta_s}$ with trainable parameters $\theta_s$:
\begin{equation}
    \bm{f}_s = F_{\theta_s}(\hat{M}_h, \hat{M}_s) \in \mathbb{R}^d,
\end{equation}
where $\bm{f}_s$ is the scene feature, which compactly represents a 3D scene. Consider a space-varied and scene-varied latent space, \ie given a 3D position $\bm{x} \in \mathbb{R}^3$ and a scene feature $\bm{f}_s \in \mathbb{R}^d$, we parameterize it into a hyperspace, \ie $\mathbb{R}^{3+d} \xrightarrow{} \mathbb{R}^{C_H}$, using a neural hash function with trainable encoding parameters $\theta_H \in \mathbb{R}^{L\times T\times C_H}$:
\begin{align}
\label{eq:hash-func}
    F_{\theta_H}(\bm{x}, \bm{f}_s) &= H(\bm{x}, \bm{f}_s; \theta_H), \\
    H(\bm{x}, \bm{f}_s) &= \Big(\bigoplus^d_{i=1}f_s^i\pi^i \bigoplus^3_{j=1}x^j\pi^j \Big) \mod T,
\end{align}
where $\oplus$ denotes the bit-wise XOR operation and $\pi^i, \pi^j$ are large and unique prime numbers. Note that, we construct $L$ levels multiresolution hash grids to represent multiscale features, $T$ is the maximum number of entries per level, and $C_H$ is the number of channels of each unique feature vector. 
We set $L=16, T=2^{19}, C_H=8$.
Specifically, the two inputs $(\bm{x}, \bm{f}_{s})$ are concatenated before feeding into the hash grid. As the dimension of the scene feature is set to $d=2$, therefore the input to the hash grid is five-dimensional. The unique prime numbers in Eq.~\ref{eq:hash-func} are set accordingly\footnote{$\pi^1 = 1, \pi^2 = 2654435761, \pi^3 = 805459861, \pi^4 = 3674653429, \pi^5 = 2097192037$. The first prime number is set to 1 for better cache coherence while keeping pseudo-independence~\cite{1949Mathematical} of the dimensions on the hashed value.}.

\subsection{Unbounded 3D Scene GAN in the Wild}
\label{sec:nerfgan}
In this section, we introduce our adversarial training framework which learns to generate unbounded 3D scenes from in-the-wild 2D image collections without camera parameters. The key is the semantic alignment of content throughout the whole pipeline.

\paragraph{Generator.}
The generator $G$ is a neural volumetric renderer~\cite{mildenhall_nerf_2020}, defined as 
\begin{equation}
    G(\bm{z}, \bm{z}_{\mathrm{style}}, \bm{o}, \bm{d}; \Phi_G) = \mathcal{R}(F_{\Phi_G}(\bm{z}, \bm{z}_\mathrm{style}), \bm{o}, \bm{d}),
\end{equation}
where $\bm{z} \sim p_{\mathrm{scene}}$ is sampled from simplex noise (Sec.~\ref{sec:orthogonal}) to represent different scenes, \ie $\bm{z} \xrightarrow{} (\hat{M}_h, \hat{M}_s)$. The style code $\bm{z}_{\mathrm{style}} \sim p_{\mathrm{style}}$ is sampled from normal distribution to represent different styles, \eg different weathers and illuminations within a given scene. And $\mathcal{R}$ is a volumetric renderer based on a conditional neural radiance field:
\begin{equation}
\label{eq:cond-nerf}
    F_{\Phi_G}(\bm{z}, \bm{z}_\mathrm{style} | \bm{x} \in \mathbb{R}^3) = \{\bm{c}(\bm{f}_{\bm{x}}, \bm{z}_\mathrm{style}, l(\bm{x})), \bm{\sigma}(\bm{f}_{\bm{x}})\},
\end{equation}
where $\bm{c}$ is color and $\bm{\sigma}$ is volume density. Given a 3D position $\bm{x}$ in the local volume $V_{w}$, we can obtain its semantic label $l(\bm{x})$ by sampling from $\hat{M}_s$, and the indexed latent feature $\bm{f}_{\bm{x}}$ via Eq.~\ref{eq:hash-func}. As shown in Fig.~\ref{fig:overview}, the renderer takes $\bm{f}_{\bm{x}}$ as input, and is modulated by $l(\bm{x})$ and style code $\bm{z}_{\mathrm{style}}$. Then, the output image is generated via volume rendering. Given a sampled camera view with origin at $\bm{o}$ and view direction from $\bm{d}$, we cast a ray $\bm{r}(t) = \bm{o} + t\bm{d}$, the corresponding pixel value $C(\bm{r})$ is obtained by an integral:
\begin{equation}
\label{eq:rendering}
    C(\bm{r}) = \int^{\infty}_{0} T(t)\bm{c}(\bm{f}_{\bm{r}(t)}, \bm{z}_{\mathrm{style}}, l(\bm{r}(t)))\bm{\sigma}(\bm{f}_{\bm{r}(t)})dt,
\end{equation}
where $T(t) = \mathrm{exp}(-\int^t_0\sigma(\bm{f}_{\bm{r}(s)})ds)$.

\paragraph{Discriminator.} To reinforce the semantic alignment of content, we employ a semantic-aware discriminator~\cite{schonfeld_you_2021} to discriminate between real and fake images. For each perspective camera view at $(\bm{o}, \bm{d})$, we obtain the segmentation map $\bm{S}_f$ of the generated images $\bm{I}_f = G(\bm{z}, \bm{z}_{\mathrm{style}}, \bm{o}, \bm{d}; \Phi_G)$ by accumulating semantic labels $l(\bm{x})$ sampled from the semantic field along each ray. Then, the fake pairs $(\bm{I}_f, \bm{S}_f)$ and the real pairs $(\bm{I}_r, \bm{S}_r)$ are fed into discriminator $D(\bm{I} | \bm{S}; \Phi_{D})$ for adversarial training.

\begin{figure*}[t]
    \begin{center}
    \centerline{\includegraphics[width=1.0\linewidth]{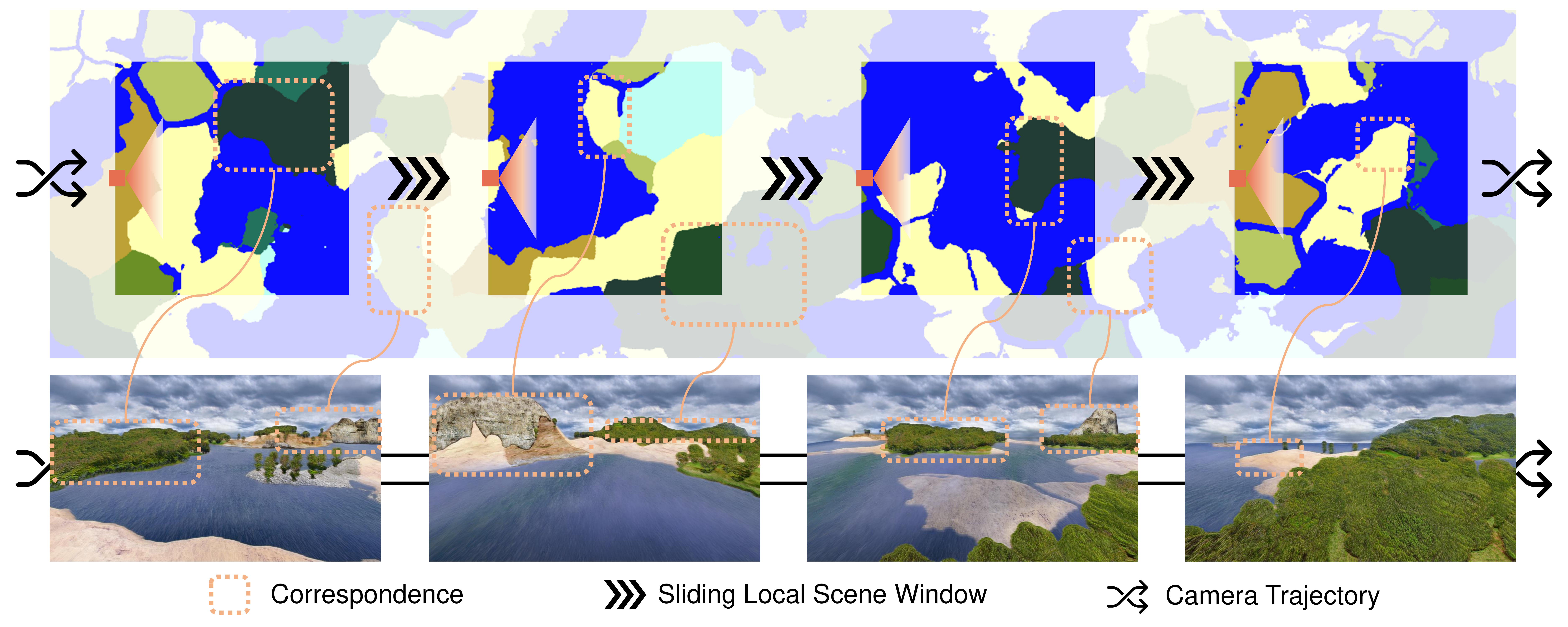}}
    \caption{\textbf{Sliding window mechanism to generate unbounded scenes beyond training resolution}. Given a scene with its \scenerepresentation\ size of $10240\times10240$, we first generate the BEV maps for the entire world, then bind the local scene window (highlighted rectangles) to the camera position (orange). Given a fly-through camera trajectory, the local scene window slides accordingly to render coherent frames (bottom).
    }\label{fig:sliding}
    \end{center}
    \vspace{-0.2in}
\end{figure*}

\begin{figure}[t]
    \begin{center}
    \centerline{\includegraphics[width=1.03\columnwidth]{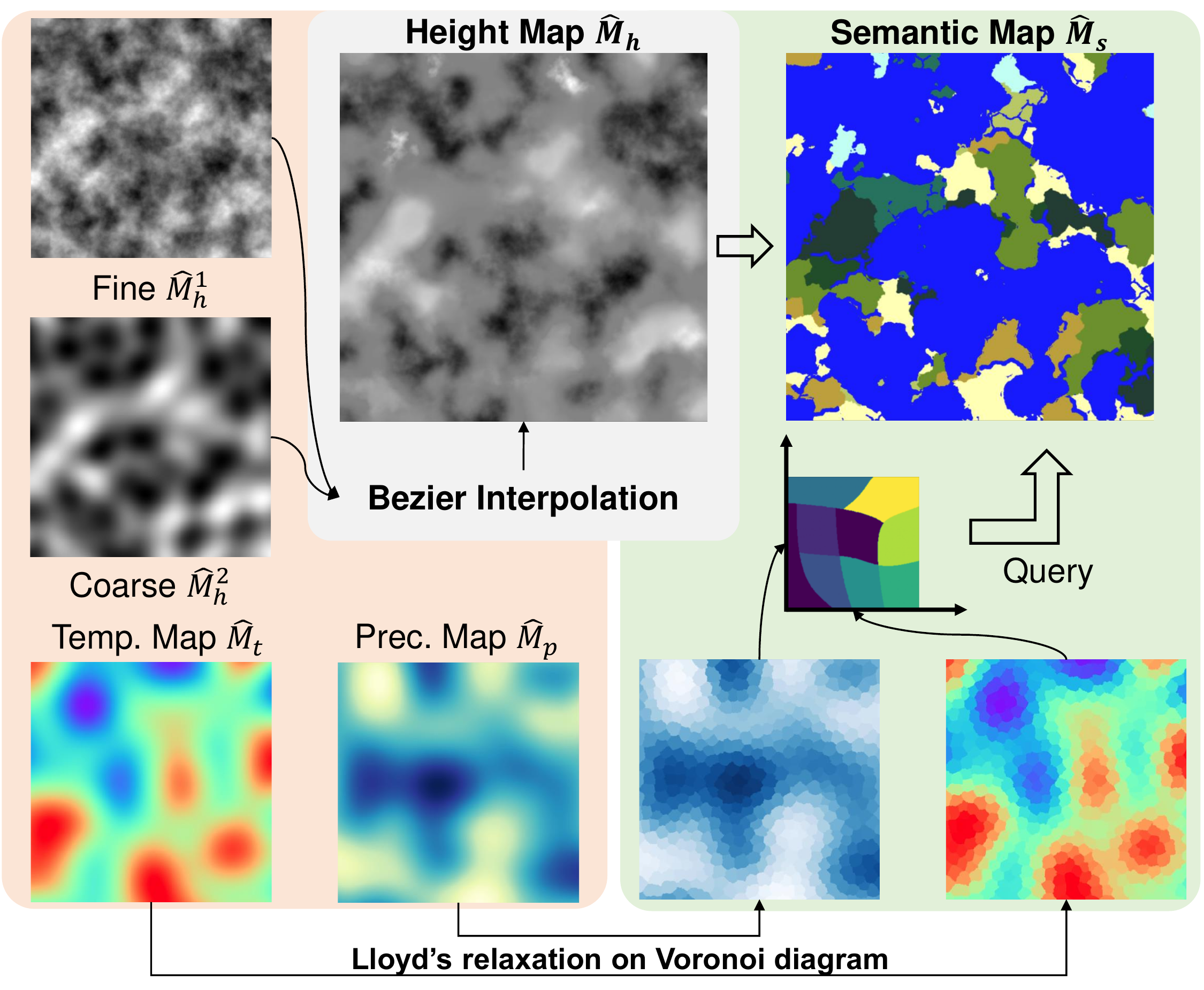}}
    \caption{\textbf{Procedural Generation of BEV Scene Representation}. The feed-forward mapping from random noise $z$ to the BEV scene representation (Sec.~\ref{sec:orthogonal}), \ie $z \xrightarrow{} (M_h, M_s)$, can be either learned from data or parameter-free. Our instantiation starts from 2D simplex noises (highlighted in \textbf{\textcolor{lightorange}{orange background}}). Please refer to Sec.~\ref{sec:pcg} for details.
    }\label{fig:pcg}
    \end{center}
    \vspace{-0.25in}
\end{figure}

\section{Implementation Details}
\label{sec:detail}

\subsection{BEV Scene Representation Generation}
\label{sec:pcg}
As introduced in Sec.~\ref{sec:orthogonal}, the feed-forward mapping from random noise $z$ to the BEV scene representation, \ie $z \xrightarrow{} (M_h, M_s)$, can be either learned from data or parameter-free. Our instantiation is a parameter-free way via procedural generation, which enables us to scale up the training data for free. We let $z$ be a simplex noise~\cite{perlin1985image, simplex} which is commonly used to generate continuous and infinite terrains in computer graphics, and generate the height field $M_h$ and the semantic field $M_s$ with any resolution respectively. 

Simplex noise~\cite{simplex} is an $n$-dimensional noise function based on gradient grids, which is an improved version of Perlin noise~\cite{perlin1985image} that has lower computational complexity and fewer directional artifacts. Specifically, we consider $z$ to be a 2D simplex noise in \framework, \ie $n = 2$. The procedure to generate the height field and the semantic field is presented in Fig.~\ref{fig:pcg}.

\begin{figure}[t]

    \begin{center}
    \centerline{\includegraphics[width=1.0\columnwidth]{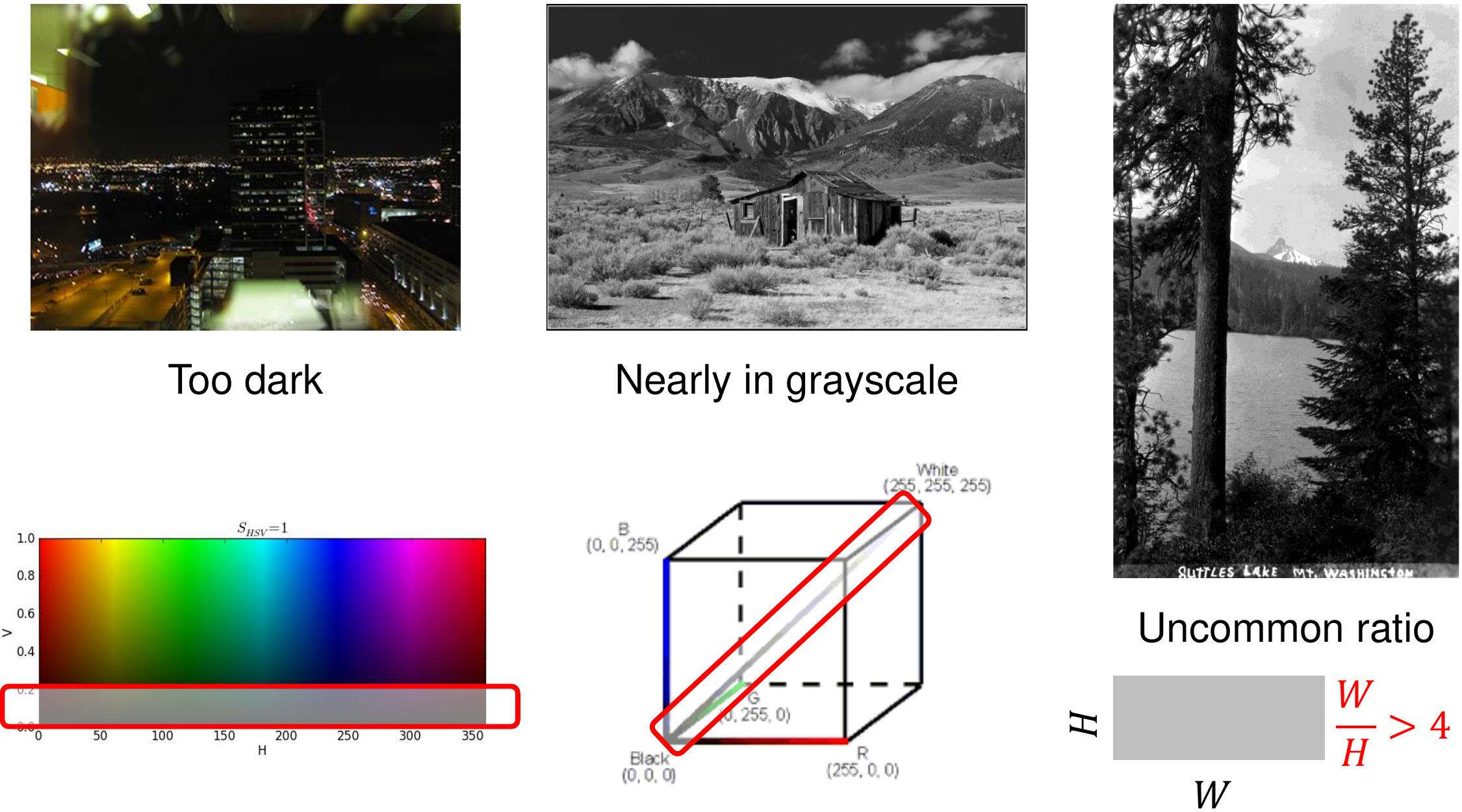}}
    \caption{\textbf{Dataset Preprocessing}. We apply rule-based filters to remove images that are too dark, nearly in grayscale, and with uncommon ratios. The filters are highlighted in \textcolor{red}{red box}. We remove dark images whose values in HSV color space are lower than the threshold. Grayscale images are identified by computing average grayscale in RGB color space, and filtered based on the L2 distance to the gray axis.
    }\label{fig:dataset}
    \end{center}
    \vspace{-0.2in}
\end{figure}

\paragraph{Height Field Generation.}
As for the height field $M_h$, we first sample two 2D simplex noise maps with a fixed random seed yet different levels of the octave as $\hat{M}^1_h, \hat{M}^2_h$. The level of the octave controls the fine-grained details in the noise map, which leads to a smoother appearance with a lower level of the octave. One can observe that a higher level of the octave leads to sufficient details but noisy heights while a lower level of the octave results in flat and smooth terrain but lacks diversity. To strike a balance in between, obtaining smooth transitions across different types of landscape, we apply bezier-based interpolation between $\hat{M}^1_h, \hat{M}^2_h$ to get the final height map $\hat{M}_h$. Note that, the height field $M_h$ is the continuous form of height map $\hat{M}_h$, which is approximated by increasing the resolution of the height map.

\begin{figure*}[t]
    \begin{center}
    \centerline{\includegraphics[width=2.0\columnwidth]{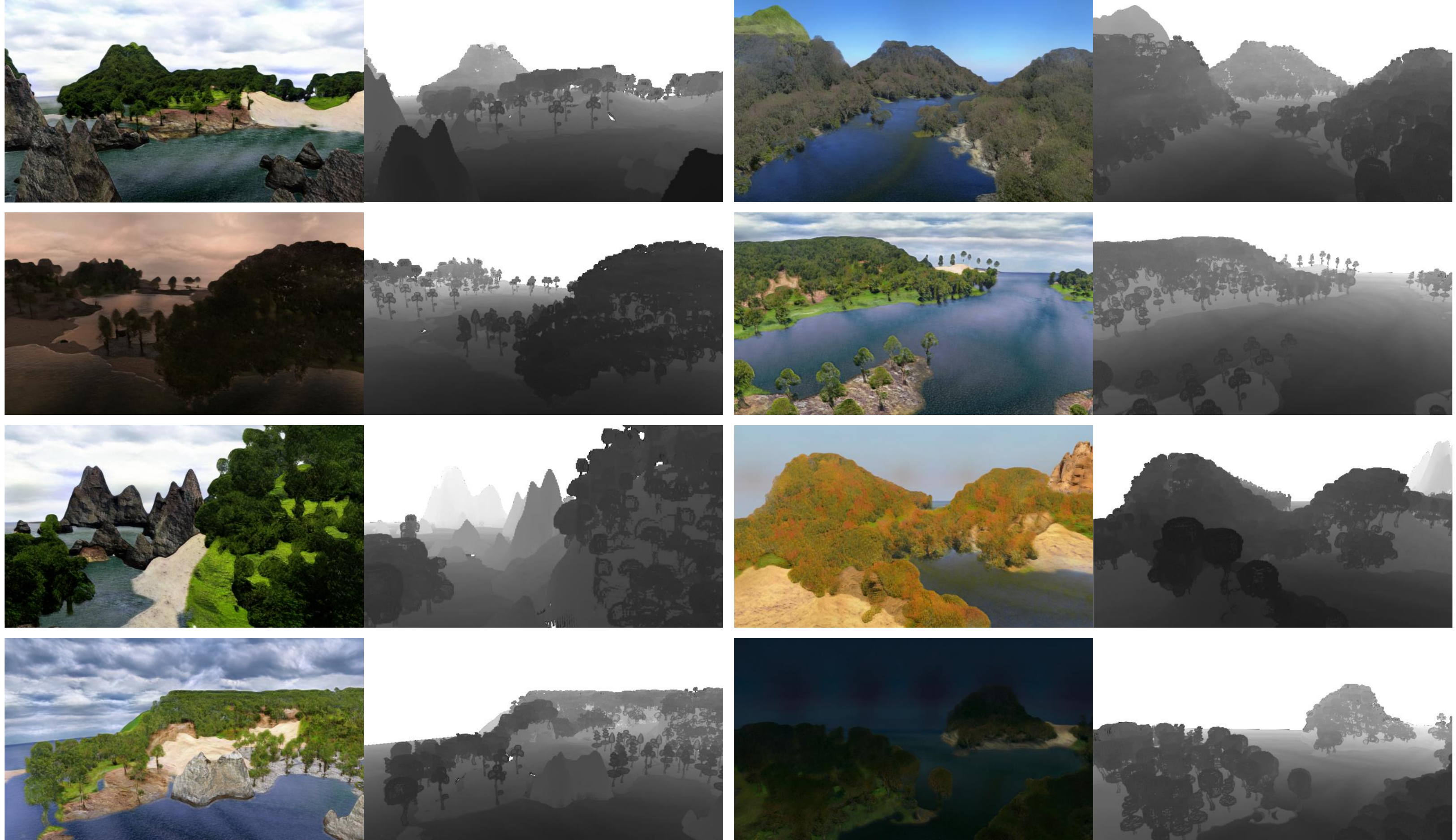}}
    \caption{\textbf{RGB renderings and corresponding depth maps of \framework}. Our method can synthesize different 3D landscapes in different styles with free camera trajectories, where both the appearance and geometry are of promising quality.}\label{fig:main-vis}
    \end{center}
    \vspace{-0.1in}
\end{figure*}

\paragraph{Semantic Field Generation.}
As for the semantic field $M_s$, we derive it from another two 2D simplex noise maps with intermediate outputs as temperature map $\hat{M}_t$ and precipitation map $\hat{M}_p$ respectively. Specifically, we sample two 2D simplex noise maps with varied seeds (to ensure they have different patterns) and the same level of the octave. The first noise map is termed as temperature map $\hat{M}_t$ which denotes the spatially varied distribution of the temperature. The second noise map is termed as precipitation map $\hat{M}_p$ which represents the level of precipitation. We are inspired by geography that temperature and precipitation are two main factors that affect the semantics of landscapes. Therefore, on top of $\hat{M}_t, \hat{M}_p$, we can define the semantics of the landscape using a 2D lookup table $I_{\mathrm{LUT}}$, which is an image with a resolution of $N_t \times N_p$ that stores the corresponding semantic label given a certain temperature and precipitation. $N_t$ and $N_p$ are the range resolution of the temperature and precipitation respectively, both of which we set to 256 in our implementation. Totally, there are 9 different landscapes in $I_{\mathrm{LUT}}$, including desert, savanna, woodland, tundra, seasonal forest, rain forest, taiga, temperate forest, and grassland.

\paragraph{Label Regularization.}
For our semantic-conditioned GAN training framework, we introduce a new set of 12 classes of semantic labels as a higher level of abstraction: sky, tree, dirt, flower, grass, gravel, water, rock, stone, sand, snow, and others. For instance, the desert landscape would be full of sand labels, the tundra landscape consists of a mixture of grass, dirt, rock, and stone, and the region with negative height is treated as water. To prevent noisy edges and unrealistic holes within different landscapes, we use the Voronoi diagram followed by Lloyd's relaxation to regularize the semantic map $\hat{M}_s$. Specifically, the semantic label of each point is quantized to most of the semantic labels within a Voronoi cell. This regularization ensures the consistency and smoothness of the semantic field.

\subsection{Training}
\paragraph{Camera pose sampling.} Different from existing 3D-aware generative models~\cite{chan_efficient_2022, EVA3D, gu2021stylenerf}, which either assume a camera pose distribution or estimate the camera pose using off-the-shelf tools, our framework supports freely sampling camera pose with 6 DoF. By constructing the local scene window $V_w$, one can explicitly sample perspective camera in $\mathrm{SE}(3)$. To further improve the sampling efficiency during training, we employ rejection sampling technique~\cite{hao_gancraft_2021} which rejects any camera pose whose rendering produces a low mean depth or a low entropy of semantic labels.

\paragraph{End-to-end learning.} \framework\ is directly learned on in-the-wild 2D images collections in an end-to-end manner. For ease of semantic alignment, besides the semantic-aware discriminator, we use off-the-shelf models~\cite{park_semantic_2019} for semantic image synthesis to generate pseudo ground truth~\cite{hao_gancraft_2021} from $\bm{S}_f$ as the reconstruction target. We train our model with a hybrid objective which consists of a reconstruction loss and an adversarial learning loss. In specific, we leverage a combination of GAN loss~\cite{hingeloss}, mean squared error (MSE) loss, and perceptual loss~\cite{perceptual_loss}, with their weights of 0.5, 10.0, and 10.0, respectively.

During training, we randomly sample scenes with the level of detail $N^D = 2048$. The ray casting and 3D points sampling is executed within the local scene window which has a resolution of $N^D_w \times N^D_w \times H^D_w$. We choose $N^D_w = 1024, H^D_w = 256$ in our implementation. In volume rendering, the resolution of generated images is $256\times 256$ with 24 points sampled along each camera ray. Totally, we have $\{\theta_s, \theta_H, \Phi_G, \Phi_D\}$ as trainable parameters in our model, which are the parameter of the scene encoder, the generative hash grid, the volumetric renderer, and the discriminator, respectively. We adopt Adam~\cite{adamoptimizer} optimizer with $\beta = (0, 0.999)$ for training. We use a learning rate of $5\times 10^{-4}$ for $\theta_s$, $1\times 10^{-4}$ for $\theta_H$ and $\Phi_G$, $4\times 10^{-4}$ for $\Phi_D$ respectively. Our model is trained on 8 NVIDIA A100 GPUs for 600k iterations with a batch size of 8, which takes approximately 80 hours to converge. The scene encoder $F_{\theta_s}$ is implemented as a convolutional network, and the dimension of the scene feature $\bm{f}_s$ is set to $d=2$.

\begin{figure*}[t]
    \begin{center}
    \centerline{\includegraphics[width=1.96\columnwidth]{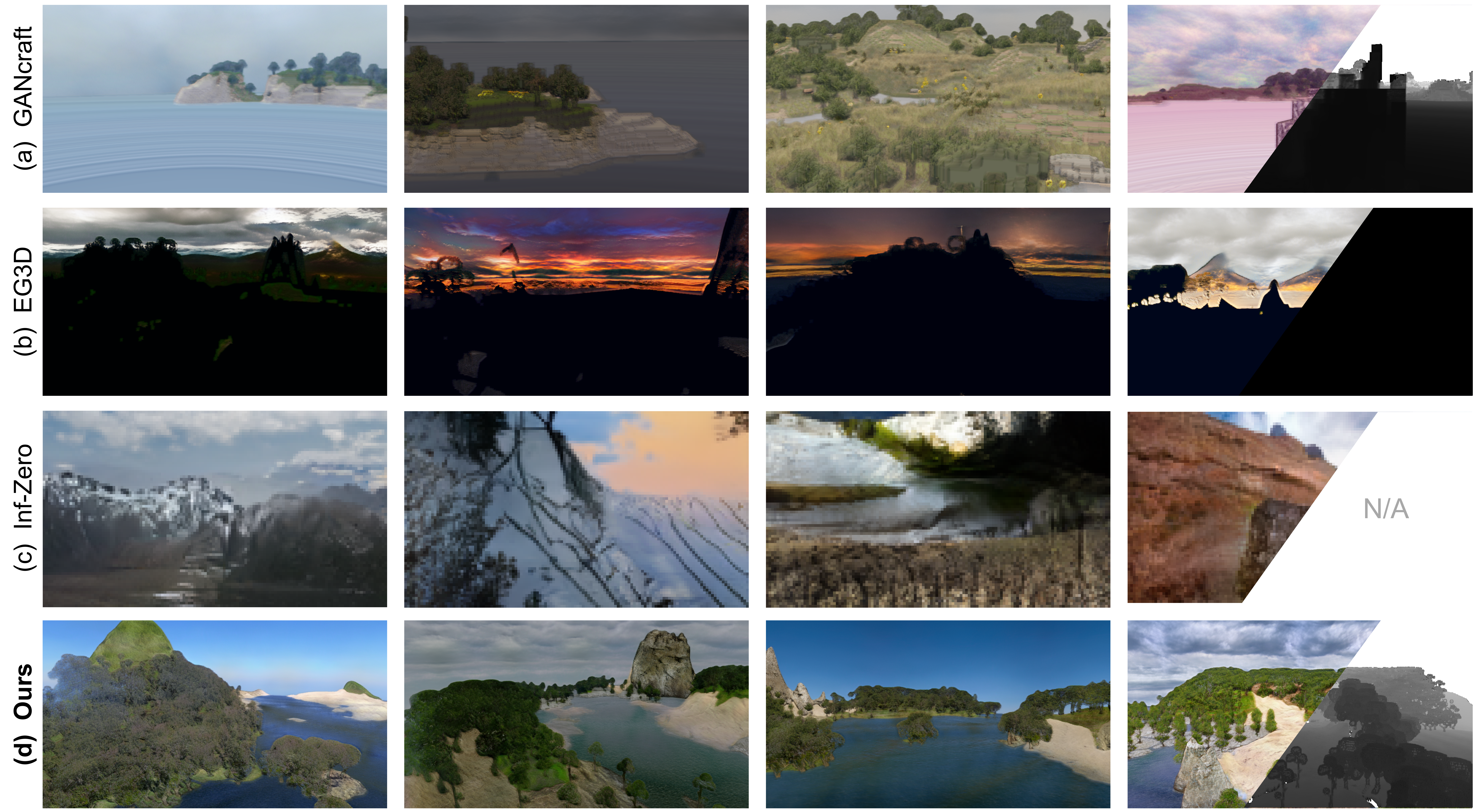}}
    \caption{\textbf{Qualitative comparison}. Both the 2D renderings and 3D depths (last column) generated by ours achieve the best visual quality. (a) GANcraft~\cite{hao_gancraft_2021}. (b) EG3D~\cite{chan_efficient_2022}. (c) Inf-Zero~\cite{li_infinitenature-zero_2022}. (d) Ours.
    }\label{fig:vis-comp}
    \end{center}
    \vspace{-0.2in}
\end{figure*}

\subsection{Inference}
\paragraph{Unbounded scene generation.} Despite training on a fixed size of local scene window, our model allows a flexible spatial size of the entire scene, enabling generating scenes in arbitrary scales. To create scenes in an unbounded domain, we employ our model in a sliding-window fashion, as illustrated in Fig.~\ref{fig:sliding}, taking a scene ($N^D = 10240$) ten times larger than the training scene size as an example. We initially generate the \scenerepresentation\ of the entire world, then bind the local scene window to the camera position. Note that, $f_s$ is the scene feature encoded by the scene encoder $F_{\theta_s}$ on the entire BEV maps, which is already seamless prior to feeding in volumetric rendered. Given a camera trajectory to be rendered, the local scene window slides according to the camera positions, extracting the scene feature $\bm{f}_s$ for each frame. For each queried 3D point during rendering, we take its global normalized coordinate instead of the coordinate within the local scene window as the 3D position $\bm{x}$. Subsequently, the camera views are rendered with continuously updated $(\bm{x}, \bm{f}_s)$ along the trajectory.

\paragraph{High-resolution renderings.}
Thanks to our expressive and compact 3D representation, our framework can render images much higher than training during testing time. We can render images beyond the training resolution by specifying the camera intrinsic and the level of detail $N^D$. For example, renderings with a resolution of $1920\times 1080$ are presented in Fig.~\ref{fig:main-vis}. Furthermore, fly-through videos in 4K ($3840\times2160$) are shown in the supplementary video. Approximately, rendering a frame with a resolution of $960\times 540$ on an A100 GPU takes 2 seconds.

%% file: sections/04_experiment.tex
\begin{table}[t]
\caption{\textbf{Quantitative results on unbounded 3D scene generation.} The top three techniques are highlighted in \textcolor{rred}{red}, \textcolor{oorange}{orange}, and \textcolor{yyellow}{yellow}, respectively. $^{\dag}$Methods adapted to fit our setting.
}\label{tab:comparisons}
\centering
\begin{tabular}{ccccccc}
\toprule
     Methods& FID\ $\downarrow$& KID\ $\downarrow$ & Depth\ $\downarrow$ & CE\ $\downarrow$ & SfM rate $\uparrow$\\
     \midrule
     $^{\dag}$GANcraft~\cite{hao_gancraft_2021}&\cellcolor{yyellow}93.73&\cellcolor{yyellow}4.82&\cellcolor{oorange}0.464&\cellcolor{oorange}0.058&\cellcolor{yyellow}0.450\\
     $^{\dag}$EG3D~\cite{chan_efficient_2022}&103.86&6.20&\cellcolor{yyellow}0.993&\cellcolor{yyellow}1.178&\cellcolor{oorange}0.475\\
     Inf-Nature~\cite{liu_infinite_2021}& -& - & - &1.555&0.380\\
     Inf-Zero~\cite{li_infinitenature-zero_2022}&\cellcolor{oorange}79.99&\cellcolor{rred}4.51& -&1.213&0.250\\
     \textbf{Ours}&\cellcolor{rred}76.73&\cellcolor{oorange}4.52&\cellcolor{rred}0.277&\cellcolor{rred}0.021&\cellcolor{rred}0.935\\
     \bottomrule
\end{tabular}
\vspace{-0.2in}
\end{table}

\section{Experiments}
\label{sec:exp}

\begin{figure*}[t]
    \begin{center}
    \centerline{\includegraphics[width=2.0\columnwidth]{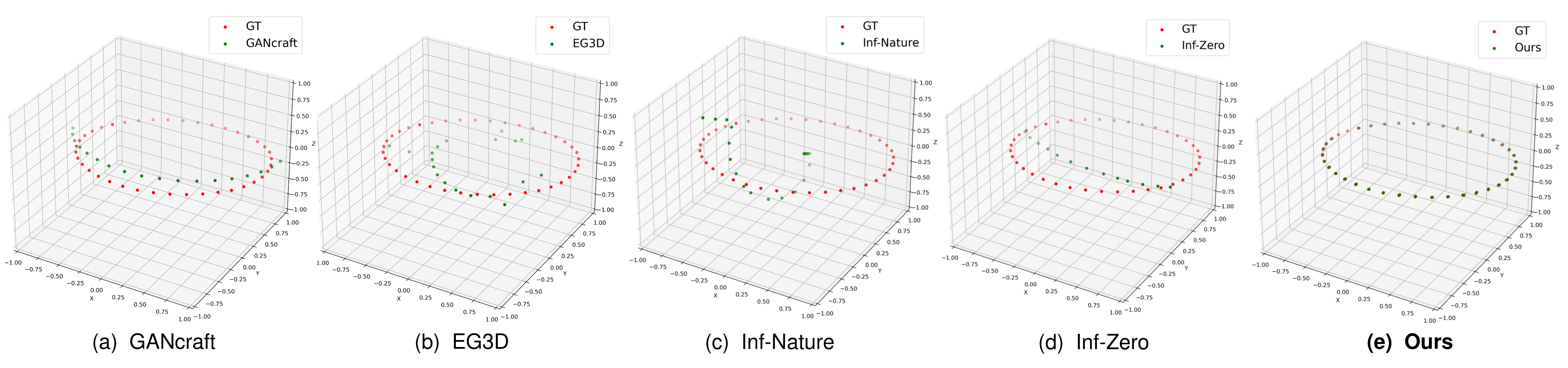}}
    \caption{\textbf{Visualization of camera trajectory for CE and SfM rate evaluation}. The red trajectory is the ground truth camera trajectory defined as Eq.~\ref{eq:circle}. And the green trajectory is successfully recovered camera poses by each method using structure-from-motion (SfM). The camera error (CE) is computed as the scale-invariant L2 distance between estimated cameras and ground truth. SfM rate is defined as the proportion of successfully recovered camera poses relative to the total number of frames. (a) GANcraft~\cite{hao_gancraft_2021}. (b) EG3D~\cite{chan_efficient_2022}. (c) Inf-Nature~\cite{liu_infinite_2021}. (d) Inf-Zero~\cite{li_infinitenature-zero_2022}. (e) Ours.
    }\label{fig:vis-cam-traj}
    \end{center}
    \vspace{-0.1in}
\end{figure*}

\subsection{Datasets}
To enable learning from rich and diverse in-the-wild scene-level content, we collect a large-scale dataset of 1,135,662 natural images from the Internet\footnote{\url{https://flickr.com/}} with keywords as ``landscape". 
However, the raw dataset is too noisy to support scene-level 3D-aware adversarial learning. Therefore, we use a rule-based filter to remove images that are too dark, with uncommon ratios, or in grayscale. The details of data preprocessing are presented in Fig.~\ref{fig:dataset}.

The short side of images is rescaled to 896 while keeping the original ratio. 
To obtain real pairs $(\bm{I}_r, \bm{S}_r)$ for the discriminator, we extract the corresponding segmentation map for each image using ViT-Adapter\footnote{\url{https://github.com/czczup/ViT-Adapter}}\cite{chen2022vitadapter} for adversarial training. We randomly set apart 15,000 images for evaluation.

\subsection{Evaluation Protocols}
During the evaluation, we sample 1024 different scenes by randomizing the simplex noise $z$. For each scene, we sample 20 different styles by randomizing the style code $z_{\mathrm{style}}$. Each sample is a fly-through video with 40 frames (with a resolution of $960\times 540$) with any possible camera trajectory. Then, we randomly select frames from those video sequences for evaluation. The evaluation metrics are listed as follows.

\paragraph{FID and KID.} Fréchet Inception Distance (FID)~\cite{NIPS2017_fid} and Kernel Inception Distance (KID)~\cite{binkowski2018demystifying} are metrics for the quality of generated images. We use the publicly available tool\footnote{\url{https://github.com/toshas/torch-fidelity}} to compute FID and KID between 15,000 generated frames and our evaluation set.

\paragraph{Depth.} We follow a similar practice in EG3D~\cite{chan_efficient_2022} for the evaluation of 3D geometry. We use a pre-trained model\footnote{\url{https://github.com/isl-org/MiDaS}}\cite{Ranftl2022} for monocular depth estimation to generate a pseudo ground truth depth map for each generated frame. The predicted depth map is generated in volume rendering by accumulating density $\bm{\sigma}$. Then, the ``ground truth" depth and the predicted depth are normalized to zero mean and unit variance to remove scale ambiguity. Finally, the depth error is computed as the L2 distance between the two. We compute depth error on 100 frames for each method.

\paragraph{CE and SfM rate.} The camera error (CE) is raised to evaluate the multi-view consistency of renderings by computing the camera position error between the inference camera trajectory and the estimated camera trajectory by a good structure-from-motion (SfM) algorithm. We use COLMAP~\cite{schoenberger2016sfm} to perform SfM to estimate the camera pose from the rendered sequence. Note that estimation errors of the camera pose occur when a video sequence is inconsistent in a 3D space, and some frames with severe pose errors are discarded by SfM due to incoherent 3D corresponding points. 
We choose a circular trajectory (shown in the supplementary video) for computing CE. Specifically, given a local scene window with a resolution of $N^D_w \times N^D_w \times H^D_w$, the camera position is sampled from the circle defined as 
\begin{equation}
\label{eq:circle}
    (x, y, z) \;\; \mathrm{s.t.}
    \begin{cases}
        (x - 0.5N^D_w)^2 + (y - 0.5N^D_w)^2 = (0.4N^D_w)^2\\
        z = 0.2 H^D_w.
    \end{cases}
\end{equation}
The rotation of the camera is set to keep the camera looking at the center. This ground truth camera trajectory is showcased in Fig.~\ref{fig:vis-cam-traj}. The camera error (CE) is computed as the scale-invariant normalized L2 distance between reconstructed camera poses and generated poses. Moreover, we select the success rate of structure-from-motion (SfM rate) as an additional measure of 3D consistency, defined as the proportion of successfully recovered camera poses relative to the total frame count.

\subsection{Comparisons}

We compare \framework\ against four methods:

\paragraph{GANcraft}~\cite{hao_gancraft_2021} is a 3D world-to-world translation model that aims to synthesize multi-view-consistent photorealistic images given 3D semantic voxels. It needs to train on a per-world basis for 4 days with 8 V100 GPUs,  which means that it cannot generalize across scenes and does not support feed-forward scene generation. That is, given a Minecraft world of a new scene, GANcraft has to re-train the 3D translation model again. For a fair comparison that enables GANcraft to get rid of per-scene optimization, we need to modify the official code\footnote{\url{https://github.com/NVlabs/imaginaire}}. In specific, to implement the adapted GANcraft (Table~\ref{tab:comparisons}), we replace the voxel-aligned features with a feature codebook conditioned on the semantic label of the voxel. It enables the adapted baseline to optimize scene-agnostic latent features that generalize across scenes. 
    
\paragraph{EG3D}~\cite{chan_efficient_2022} is a 3D-aware GAN for faces that incorporates a hybrid tri-plane representation to enable efficient GAN training. As it requires an estimated camera pose distribution as input, which does not fit our setting (unbounded scenes and free camera poses), we adapt the official code\footnote{\url{https://github.com/NVlabs/eg3d}} and only utilize the tri-plane representation. Specifically, we use random noise vector as scene feature $\bm{f}_{s}$ and replace the generative hash grid with a StyleGAN2~\cite{karras_analyzing_2020} backbone which outputs tri-planes. The latent feature $\bm{f}_{\bm{x}}$ is computed from tri-planes via interpolation. Both the rendering and training strategies are kept the same as ours. 

\paragraph{Inf-Zero}~\cite{li_infinitenature-zero_2022} aims to synthesize perpetual views given a single image as input. It enables learning from in-the-wild images for 3D-aware scene-level synthesis, but does not ensure the 3D consistency and semantic consistency of foreground contents due to the lack of 3D representation. We adapt the officially released code\footnote{\url{https://github.com/google-research/google-research/tree/master/infinite_nature_zero}}. 

\paragraph{Inf-Nature}~\cite{liu_infinite_2021} addresses the same problem as Inf-Zero, but is trained on camera trajectories with known camera poses. Therefore, it is impossible to retrain in our setting (using in-the-wild 2D images without known camera poses), which makes it unfair to compare visual quality. Thus, we only compare CE and SfM rate to evaluate 3D consistency.

\begin{figure}[t]
    \begin{center}
    \centerline{\includegraphics[width=1.0\columnwidth]{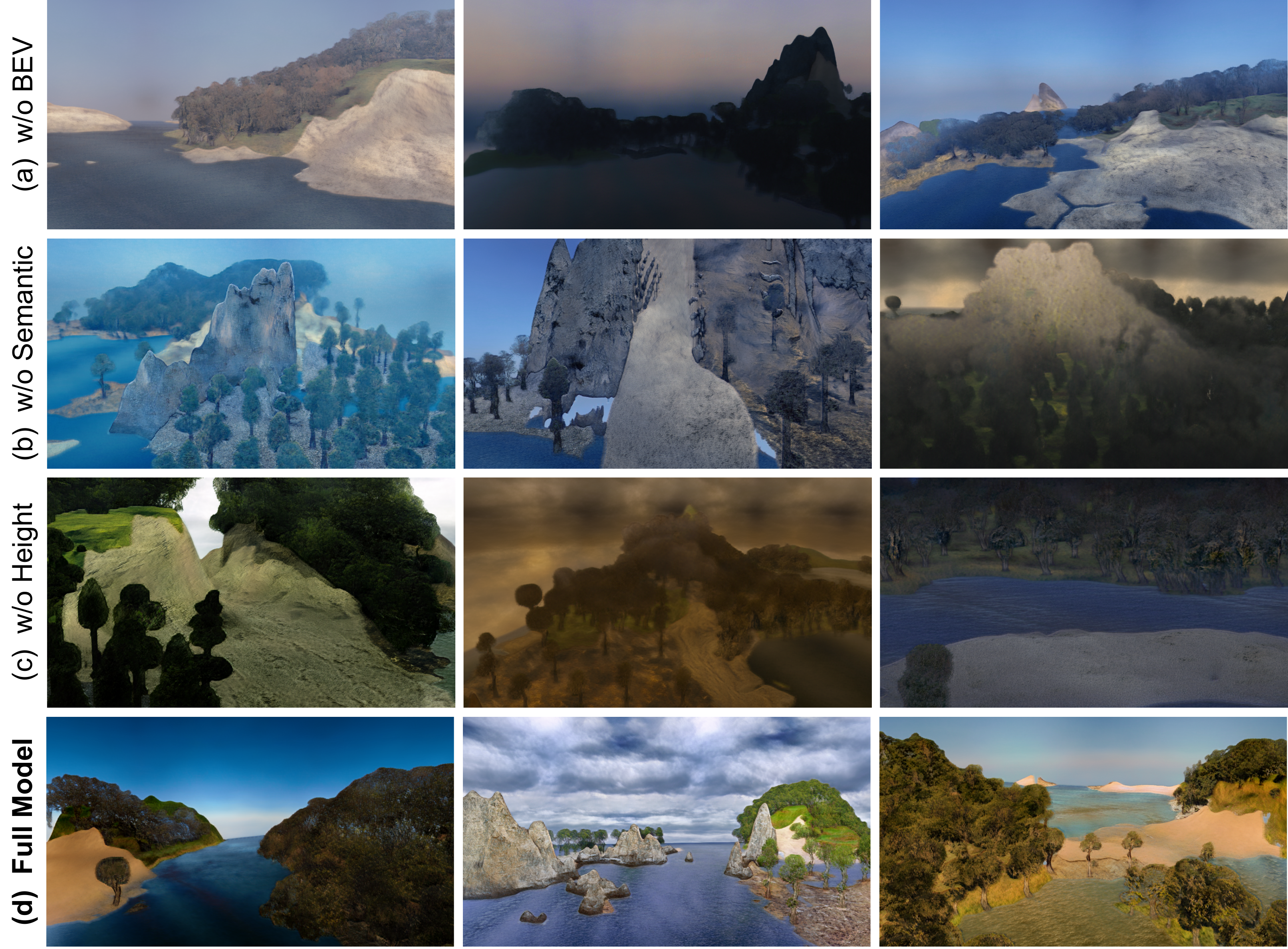}}
    \caption{\textbf{Ablation study of \scenerepresentation}. Without this representation, our model will produce dull outputs due to the lack of global semantics, which is crucial for the scene parameters to learn vivid and detailed appearance across different scenes. (a) w/o BEV: trained with no scene encoder. (b) w/o Semantic: trained with no semantic field as input. (c) w/o Height: trained with no height field as input. (d) Ours.
    }\label{fig:abl-bev}
    \end{center}
    \vspace{-0.1in}
\end{figure}

\paragraph{Qualitative Results.}
Fig.~\ref{fig:vis-comp} provides qualitative comparisons against baselines. While GANcraft synthesizes high-quality images, reliance on coarse voxels produces blocky geometry. Plus, the way of parameterizing large-scale scenes using voxel-aligned features prohibits its generalizability across different scenes, resulting in artifacts. The adapted EG3D struggles with realistic renderings due to the incapability of tri-plane for parameterizing large-scale space. Inf-Zero cannot synthesize high-resolution content with 3D consistency. Without any 3D representation, it cannot produce depth maps along with the renderings (shown as N/A). Our method generates not only images that are more view-consistent and realistic across different scenes but also plausible 3D depth. Please refer to the supplementary video for more visual results and comparisons.

\paragraph{Quantitative Results.}
Table~\ref{tab:comparisons} presents quantitative metrics of the proposed approach against baselines. Our method demonstrates significant improvements in FID and KID, which is consistent with visual comparisons. Although Inf-Zero achieves comparable image quality as ours, it is extremely 3D inconsistent due to the lack of 3D representation (indicated by the high CE and low SfM rate). High CE and low SfM rate are also observed for Inf-Nature, which validates the vital role of incorporating 3D representation in 3D scene generation. In contrast, \framework\ shows the ability to keep the correctness of 3D geometry and view consistency when producing photorealistic images, which is indicated by the lowest error in depth and CE compared with baselines.
Furthermore, we visualize the camera trajectory for evaluating CE and SfM rate in Fig.~\ref{fig:vis-cam-traj}. It is obvious that our method succeeds in recovering most of the camera poses with low errors.

\subsection{Ablation Study}

\paragraph{Effectiveness of \scenerepresentation.}
Without \scenerepresentation, the model will not have a sense of the global semantics of the entire scene within the local scene window. We measure this effect by training our model without the scene encoder $F_{\theta_s}$ and scene feature $\bm{f}_s$. Quantitative results are presented in Table~\ref{tab:abl-bev}, where both the image quality and 3D consistency are dropped. In addition, as shown in Fig.~\ref{fig:abl-bev}, the global semantics is key to ensuring vivid and realistic results by our full model (bottom row). Without this representation fed into the hash grid, the model will produce dull and unrealistic outputs (top row). Moreover, the effect of the semantic field and height field is ablated respectively. For example, ``w/o semantic'' means the model trained without semantic field as input to the scene encoder $F_{\theta_s}$. Both of them benefit the visual quality, leading to improvements in FID and KID. 

Besides, we report the efficiency of \scenerepresentation\ in Table~\ref{tab:mem-bev}. With a set GPU memory budget of 40GB, we record the highest achievable level of detail (LOD), which corresponds to the maximum size of the local scene window. The two alternatives, "Voxels" and "Semantic Voxels," explicitly represent the scene as occupied 3D voxels, exhibiting cubic complexity. The ``Voxels'' baseline stores per-voxel latent features which cannot generalize across scenes while ``Semantic Voxels'' is the representation used in the adapted GANcraft baseline. The quadratic complexity of \scenerepresentation\ significantly boosts our capacity in modeling larger scenes.

\begin{table}[t]
\caption{\textbf{Effectiveness of \scenerepresentation.} The top three techniques are highlighted in \textcolor{rred}{red}, \textcolor{oorange}{orange}, and \textcolor{yyellow}{yellow}, respectively.
}\label{tab:abl-bev}
\centering
\begin{tabular}{cccccc}
\toprule
     Methods& FID\ $\downarrow$& KID\ $\downarrow$ & Depth\ $\downarrow$  & CE\ $\downarrow$ & SfM rate\ $\uparrow$ \\
     \midrule
     w/o BEV&\cellcolor{yyellow}80.62&\cellcolor{yyellow}4.68&\cellcolor{oorange}0.314&\cellcolor{oorange}0.044&\cellcolor{oorange}0.930\\
     w/o semantic&83.20&4.77&0.401&\cellcolor{yyellow}0.046&\cellcolor{yyellow}0.910\\
     w/o height&\cellcolor{oorange}79.80&\cellcolor{oorange}4.61&\cellcolor{yyellow}0.356&0.051&0.890\\
     \textbf{Ours}&\cellcolor{rred}76.73&\cellcolor{rred}4.52&\cellcolor{rred}0.277&\cellcolor{rred}0.021&\cellcolor{rred}0.935\\
     \bottomrule
\end{tabular}
\vspace{-0.1in}
\end{table}

\begin{table}[t]
\caption{\textbf{Memory efficiency of \scenerepresentation.} 
}\label{tab:mem-bev}
\centering
\begin{tabular}{cccc}
\toprule
     Methods & Max. Supported LOD\ $\uparrow$ & FID\ $\downarrow$ & KID\ $\downarrow$ \\
     \midrule
     Voxels &$512\times512\times256$&-&-\\
     Semantic Voxels &$512\times512\times256$&93.73&4.82\\
     BEV (\textbf{Ours}) & $2048\times2048\times256$ &76.73&4.52\\
     \bottomrule
\end{tabular}
\vspace{-0.1in}
\end{table}

\begin{figure}[t]
    \begin{center}
    \centerline{\includegraphics[width=1.0\columnwidth]{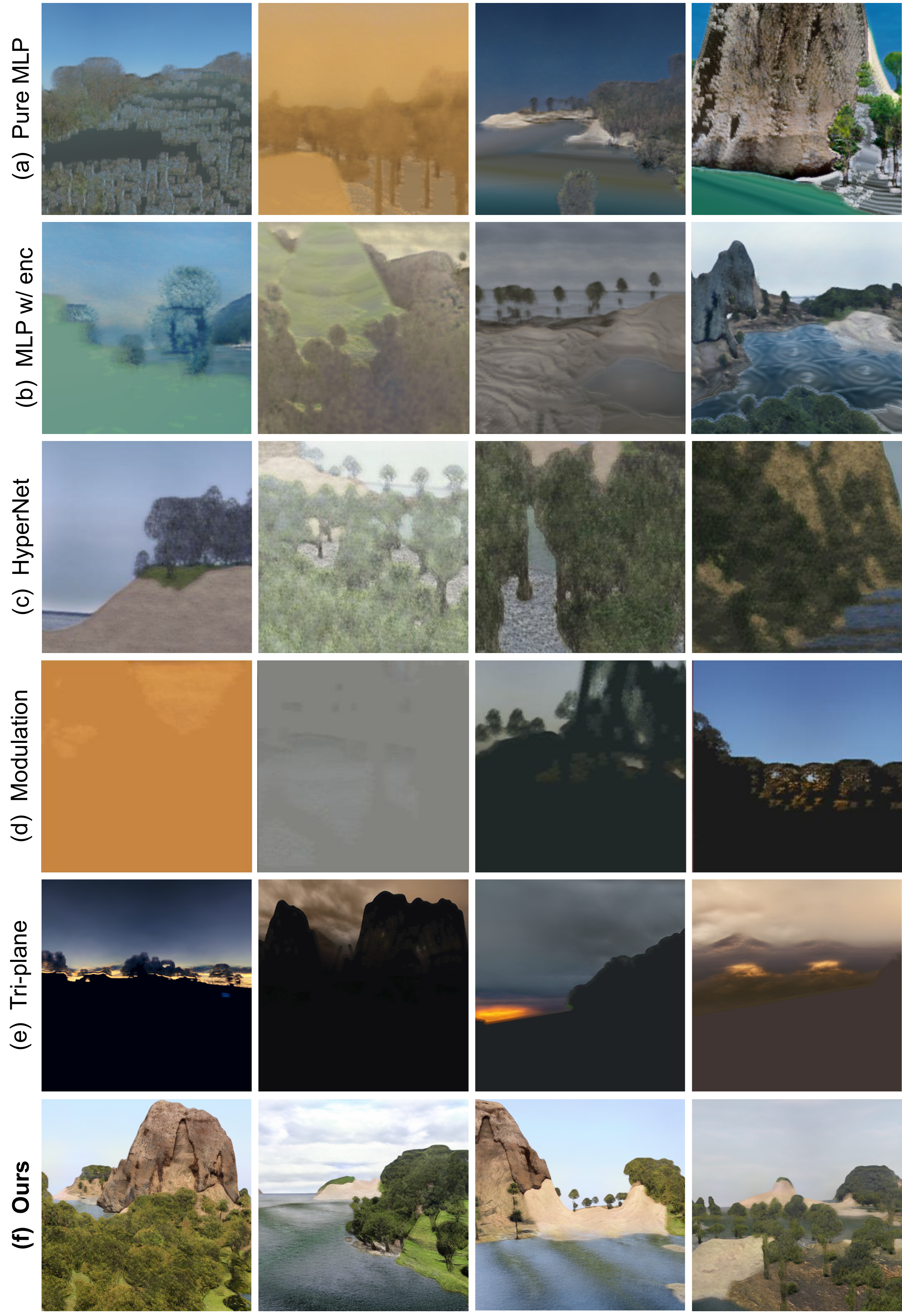}}
    \caption{\textbf{Ablation study of different generative scene parameterization}. Instead of the generative hash grid in our model, the function $F_{\theta_H}: (\bm{x}, \bm{f}_s) \xrightarrow{} \bm{f}_{\bm{x}}$ can be replaced with other parameterization techniques. Please refer to Fig.~\ref{fig:abl-hash-illu} for details of each method.
    }\label{fig:abl-hash}
    \end{center}
\end{figure}

\begin{figure}[t]
    \begin{center}
    \centerline{\includegraphics[width=1.0\columnwidth]{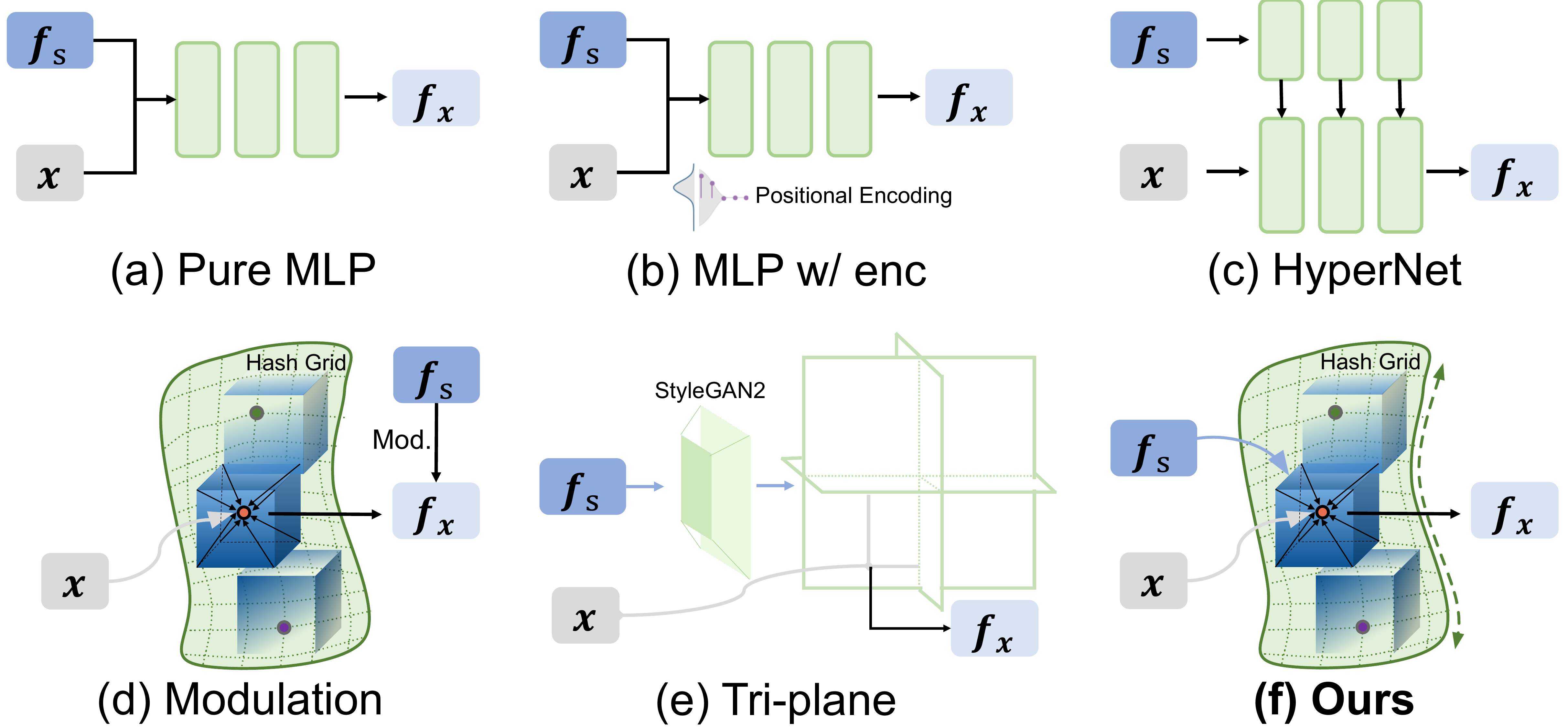}}
    \caption{\textbf{Different design choices of generative scene parameterization, \ie $F_{\theta_H}: (\bm{x}, \bm{f}_{s}) \xrightarrow{} \bm{f}_x $}. 
    }\label{fig:abl-hash-illu}
    \end{center}
    \vspace{-0.2in}
\end{figure}

\paragraph{Effectiveness of Generative Hash Grid.}
We argue that scene parameterization is also critical to the success of unbounded 3D scene generation. We ablate different ways of parameterization, \ie how to implement $F_{\theta_H}: (\bm{x}, \bm{f}_{s}) \xrightarrow{} \bm{f}_x $. We illustrated these implementations in Fig.~\ref{fig:abl-hash-illu}. We compare all five variants with our generative neural hash grid, presenting the results in Table~\ref{tab:abl-hash}. None of the methods succeeds in learning to synthesize unbounded 3D scenes from 2D images, indicated by poor FID and depth metrics. Furthermore, qualitative results are shown in Fig.~\ref{fig:abl-hash}, where our full model synthesizes vivid and appealing images. Using pure MLP leads to blocky artifacts while applying positional encoding~\cite{mildenhall_nerf_2020} produces sinusoidal patterns on the scene surface, as shown in Fig.~\ref{fig:abl-hash}(a, b). ``HyperNet'' and ``Modulation'' baselines fail to converge, yielding dull appearances. And tri-plane-based method struggles to synthesize meaningful foreground textures. We attribute the failure of these baselines to the incapability of sparsely and compactly parameterizing the scene. In terms of unbounded scenes, the observation space (image captured by camera view) always occupies a small portion of the entire scene. Thus, the parameterization should be sparsely activated, \ie only the parameter space that is related to the rendered view is sampled. In contrast, these methods model the scene as a whole, injecting instabilities in adversarial training.

\begin{table}[t]
\caption{\textbf{Effectiveness of different generative scene parameterization.} The top three techniques are highlighted in \textcolor{rred}{red}, \textcolor{oorange}{orange}, and \textcolor{yyellow}{yellow}.
}\label{tab:abl-hash}
\centering
\begin{tabular}{cccccc}
\toprule
     Methods& FID\ $\downarrow$& KID\ $\downarrow$ & Depth\ $\downarrow$ & CE\ $\downarrow$ & SfM rate\ $\uparrow$\\
     \midrule
     Pure MLP&\cellcolor{yyellow}88.76&\cellcolor{yyellow}5.07&\cellcolor{oorange}0.433&\cellcolor{yyellow}0.194&\cellcolor{yyellow}0.650\\

     MLP w/ enc&\cellcolor{oorange}83.14&\cellcolor{oorange}4.70&\cellcolor{yyellow}0.522&\cellcolor{oorange}0.106&\cellcolor{oorange}0.815\\
     HyperNet &117.64&6.07&0.781&1.032&0.525\\
     Modulation & 120.88 & 6.94 & 0.994& 1.331&0.325\\
     Tri-plane&103.86&6.20&0.993&1.178&0.475\\
     \textbf{Ours}&\cellcolor{rred}76.73&\cellcolor{rred}4.52&\cellcolor{rred}0.277 &\cellcolor{rred}0.021&\cellcolor{rred}0.935\\
     \bottomrule
\end{tabular}
\vspace{-0.1in}
\end{table}

\begin{figure}[t]
    \begin{center}
    \centerline{\includegraphics[width=1.0\columnwidth]{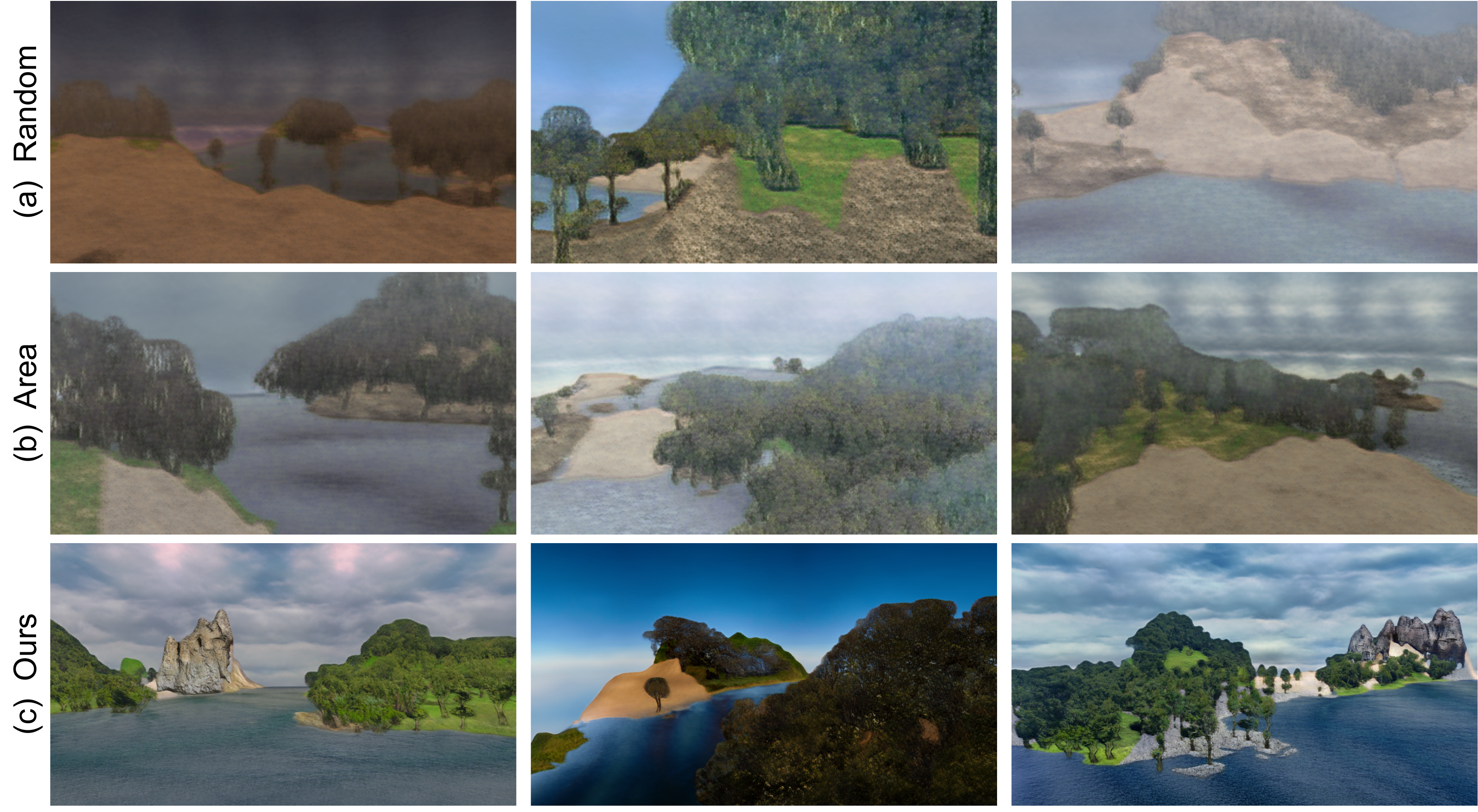}}
    \caption{\textbf{Ablation study of different camera sampling strategies during training}. (a) Random: sample camera pose in $\mathrm{SE}(3)$. (b) Area: assume cameras are distributed within a height interval. (c) Ours: rejection sampling based on mean depth and entropy of camera view.
    }\label{fig:abl-cam-traj}
    \end{center}
\end{figure}

\begin{table}[t]
\caption{\textbf{Ablation of different camera sampling strategy.}
}\label{tab:cam_sample}
\centering
\begin{tabular}{cccccc}
\toprule
     Methods& FID\ $\downarrow$& KID\ $\downarrow$ & Depth\ $\downarrow$ & CE\ $\downarrow$ & SfM rate\ $\uparrow$\\
     \midrule
     Random&102.60&6.30&0.456&0.093&0.833\\
     Area&96.90&6.13&0.429&0.094&0.865\\
     \textbf{Ours}&\cellcolor{rred}76.73&\cellcolor{rred}4.52&\cellcolor{rred}0.277&\cellcolor{rred}0.021&\cellcolor{rred}0.935\\
     \bottomrule
\end{tabular}
\vspace{-0.1in}
\end{table}

\paragraph{Effectiveness of Camera Sampling Strategy.} We report the effects of different camera pose sampling strategies during training in Table~\ref{tab:cam_sample}. Specifically, ``Random'' refers to the naive strategy of sampling camera poses in $\mathrm{SE}(3)$, without any constraint and assumption. ``Area'' indicates a sampling technique that assumes the training cameras are distributed within a height interval, but no rejection sampling is leveraged. Our rejection sampling strategy notably enhances rendering quality, as evidenced by the lowest FID and KID scores, as well as improved geometry correctness. Moreover, it is apparent that the two baseline sampling strategies result in blurry artifacts and washed-out patterns, as shown in Fig.~\ref{fig:abl-cam-traj}. Our sampling approach eliminates out camera views with low mean depth or low entropy of semantic labels both of which lead to the low diversity of visible semantic labels in image patches. Consequently, this benefits adversarial training as the distribution of training patches should be as semantically rich as real images, validated by the vivid appearance and low FID of our method.

\subsection{Further Analysis}
\label{sec:analysis}
\begin{figure}[t]
    \begin{center}
    \centerline{\includegraphics[width=1.0\columnwidth]{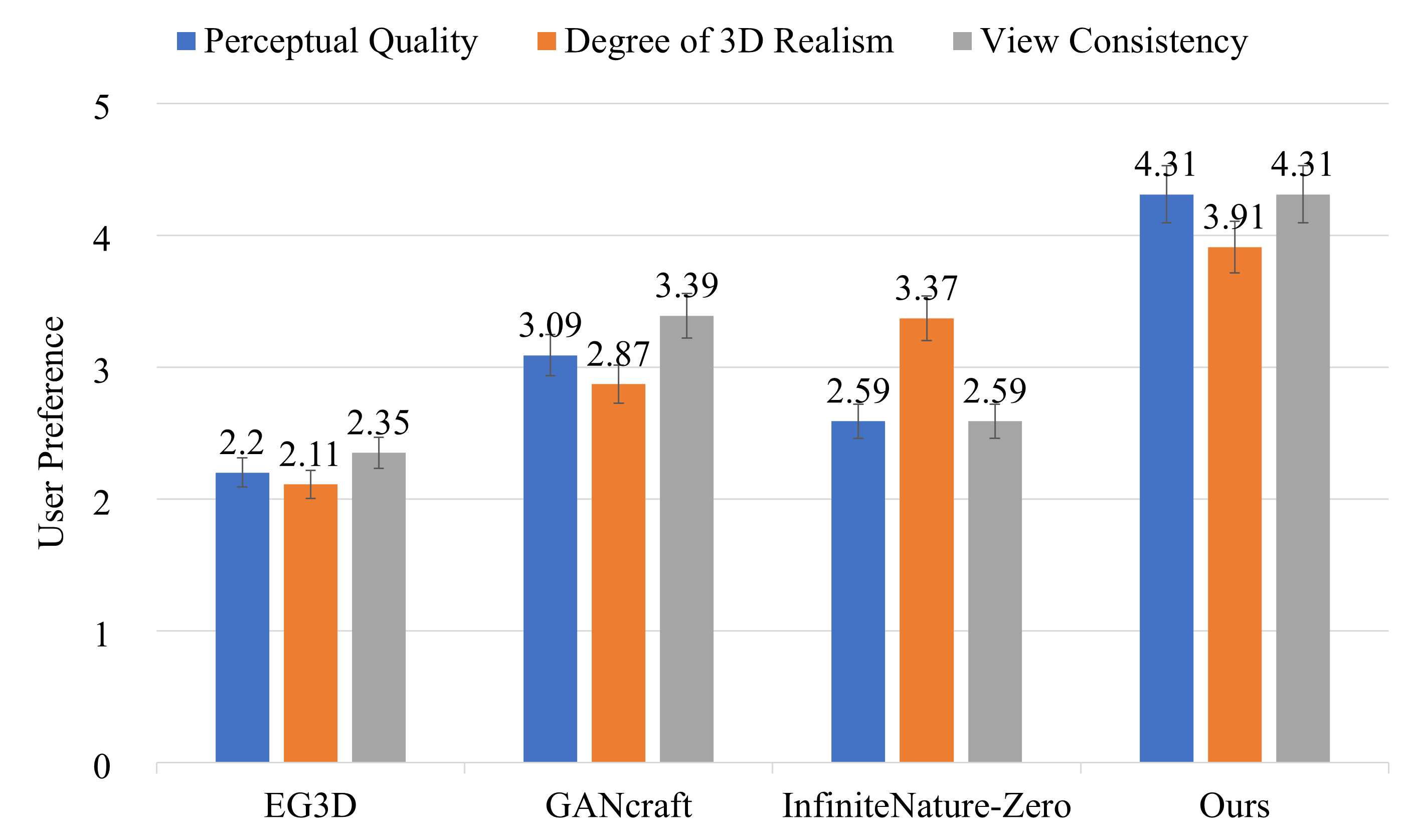}}
    \caption{\textbf{User study on unbounded 3D scene generation}. All scores are in the range of 5, and 5 indicates the best.
    }\label{fig:user-study}
    \end{center}
    \vspace{-0.2in}
\end{figure}

\paragraph{User Study.}
To better evaluate the 3D consistency and quality of unbounded 3D scene generation, we conduct an output evaluation~\cite{userstudy} as the user study, which is a survey that involves asking participants to rate alternatives. For each generated camera trajectory, a total of 20 volunteers are asked to score: \textbf{1)} the perceptual quality of the imagery, \textbf{2)} the degree of 3D realism, and \textbf{3)} 3D view consistency. All scores are in the range of 5, and 5 indicates the best. The results are presented in Fig.~\ref{fig:user-study}. The proposed method outperforms baselines by a large margin.

\begin{figure*}[t]
    \begin{center}
    \centerline{\includegraphics[width=2.0\columnwidth]{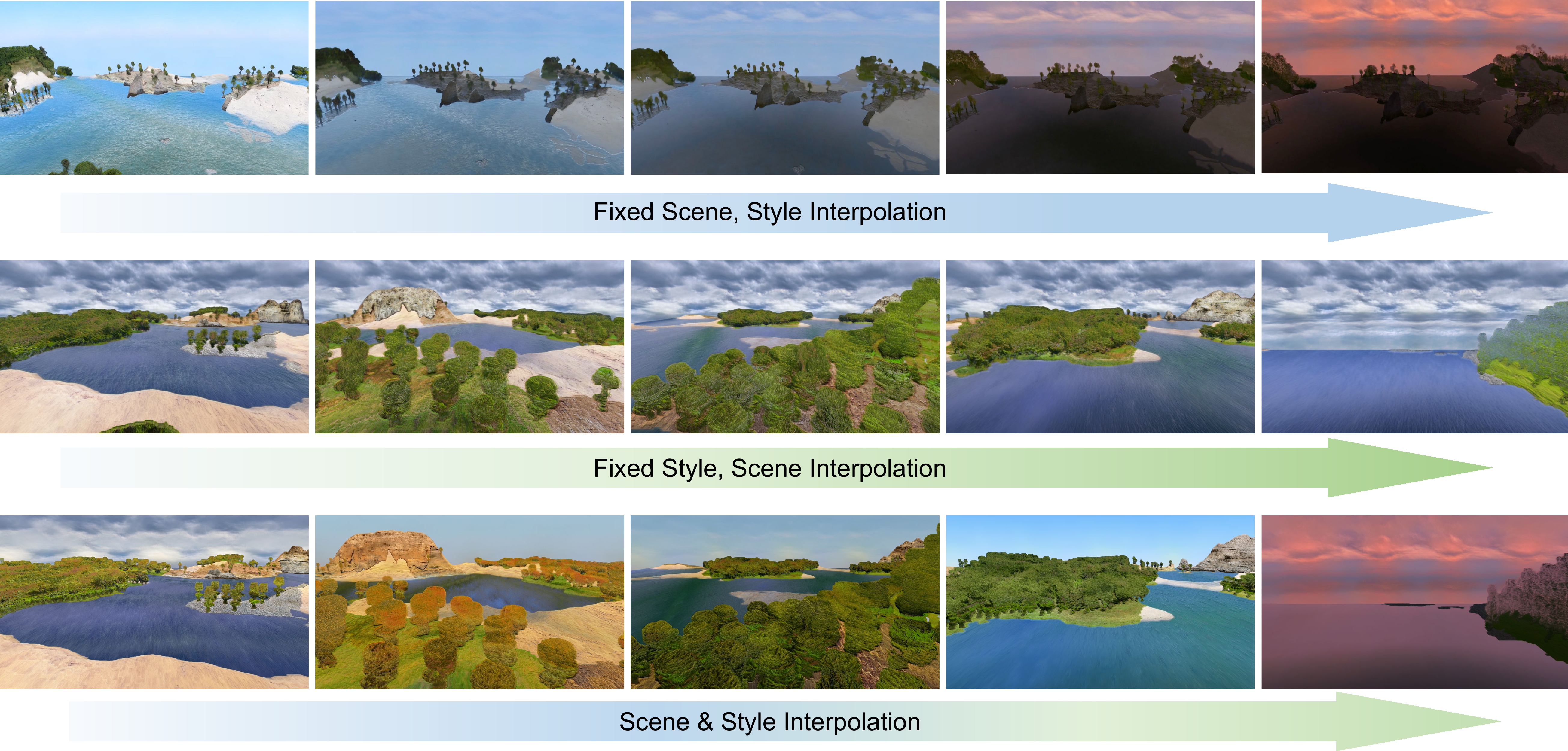}}
    \caption{Interpolation along two different dimensions, scene and style, which are controlled by $\bm{z}$ and $\bm{z}_\mathrm{style}$ respectively.
    }\label{fig:intep}
    \end{center}
    \vspace{-0.25in}
\end{figure*}

\begin{figure}[t]
    \begin{center}
    \centerline{\includegraphics[width=1.0\columnwidth]{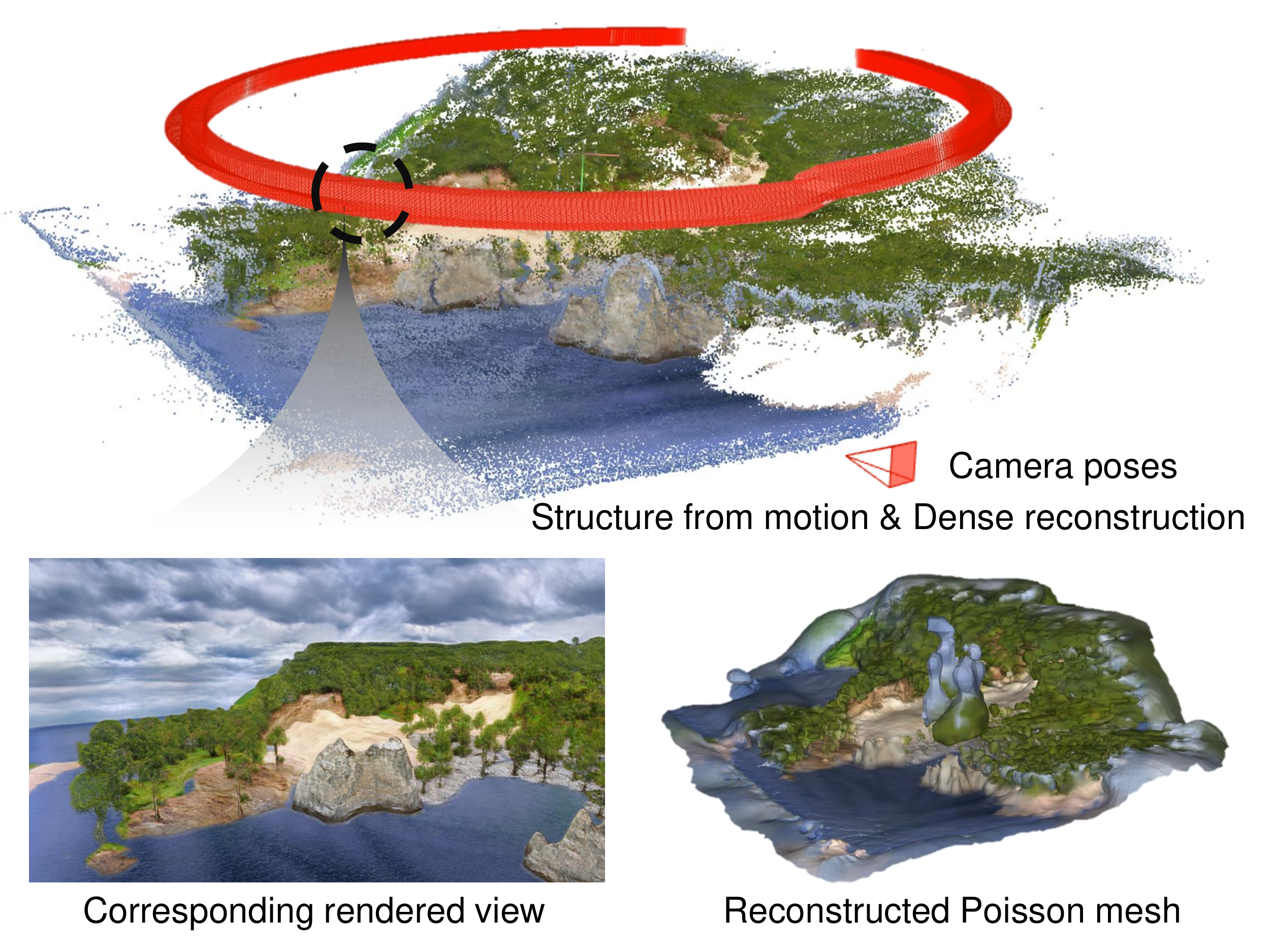}}
    \caption{\textbf{COLMAP~\cite{schoenberger2016mvs, schoenberger2016sfm} reconstruction of a 600-frame synthesized video which followed a circle trajectory}. The estimated camera poses (red) and well-defined point clouds demonstrate highly multi-view-consistent renderings of \framework.
    }\label{fig:colmap}
    \end{center}
    \vspace{-0.2in}
    
\end{figure}

\paragraph{View Consistency.}
To demonstrate the multi-view-consistent renderings of \framework, we employ COLMAP~\cite{schoenberger2016mvs, schoenberger2016sfm} to perform structure-from-motion and dense reconstruction from a synthesized video sequence. We render a video sequence of 600 frames (with a resolution of $960\times 540$), taken from a circle camera trajectory orbiting in the scene given a fixed height and looking at the center (like the sequence we show in the supplementary video). Only images are used for reconstruction, even without specifying camera parameters. As shown in Fig.~\ref{fig:colmap}, the estimated camera poses exactly match our sampled trajectory, and the resulting point cloud is well-defined and dense. Note that, comparison methods failed during dense reconstruction. It further validates that \framework\ can generate 3D consistent large-scale scenes.

\subsection{More Applications}
\label{sec:application}

\paragraph{Large-scale Landscape Generation.} The strong capability of our model in generating large-scale landscapes is validated in Fig.~\ref{fig:main-vis}. \framework\ can synthesize diverse scenes with different styles and well-defined depths, where the camera can freely move.

\paragraph{Perpetual View Generation~\cite{liu_infinite_2021}.} Given a single RGB image, the goal is to synthesize a video depicting a scene captured from a forward flying camera with an arbitrary long trajectory. Though the input is different, \framework\ can also do the task of generating videos of infinite 3D scenes unconditionally, by sliding the local scene window on an arbitrary large semantic map (Fig.~\ref{fig:sliding}). Please refer to the supplementary for our generated video with long trajectories.

\paragraph{Interpolation.} As shown in Fig.~\ref{fig:intep}, we are able to interpolate along two different dimensions, scene and style, which are controlled by $\bm{z}$ and $\bm{z}_\mathrm{style}$ respectively. The first row demonstrates a linear interpolation between two style codes to generate a smooth transitional path within the latent space, highlighting that the learned style space is semantically meaningful. The second row shows that we can interpolate between scenes to smoothly change the scene configuration while keeping the style unchanged. The bottom row showcases when the scene code and style code are interpolated simultaneously.

\begin{figure}[t]
    \begin{center}
    \centerline{\includegraphics[width=1.0\columnwidth]{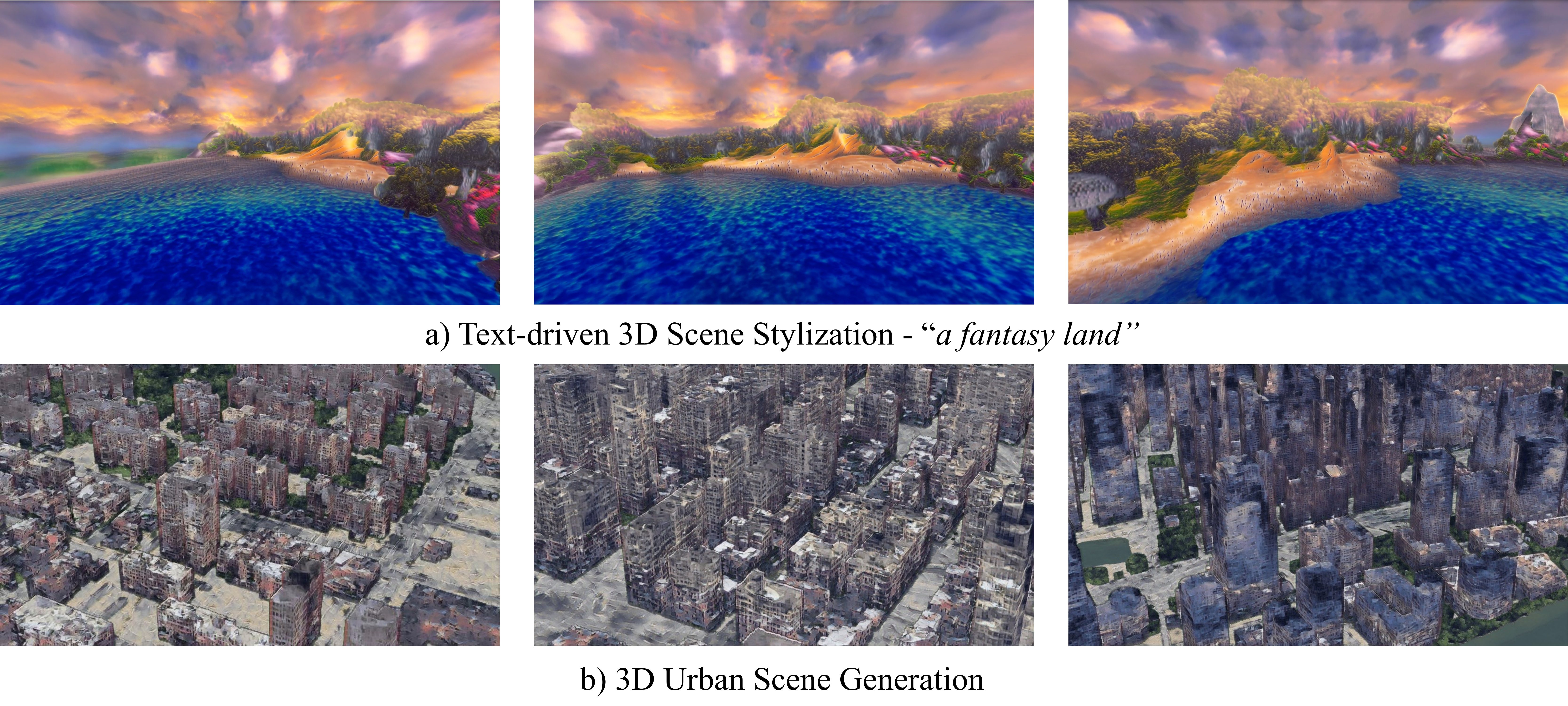}}
    \caption{\textbf{Scalability of \framework\ to other scene generative tasks.} We demonstrate the scalability of our work in two ways: \textbf{a)} text-driven 3D scene stylization and \textbf{b)} 3D urban scene generation. We only change the training data and configuration to train these models without significant modifications to our framework.}\label{fig:potential}
    \end{center}
    \vspace{-0.1in}
    
\end{figure}

\paragraph{Scalability to other generative tasks for 3D scenes.} As shown in Fig.~\ref{fig:potential}, we can easily extend \framework\ to other large-scale scene generative tasks with promising results by changing the training data and configuration only. For text-driven 3D scene stylization, as shown in Fig.~\ref{fig:potential}(a), we fine-tune our model on the images generated by ControlNet~\cite{zhang2023adding}. For urban scene generation shown in Fig.~\ref{fig:potential}(b), we train our model on image collections of urban scenery. We hope this will shed light on future work regarding 3D generative models for unbounded scenes.

%% file: sections/05_conclusion.tex
\section{Discussion}
\label{sec:conclusion}
To conclude, we present \framework\ for unbounded 3D scene generation, which aims to synthesize large-scale 3D landscapes. We propose an efficient yet expressive \scenerepresentation\ to model large-scale 3D scenes. Besides, we propose a novel generative neural hash grid to parameterize space-varied and scene-varied latent features, aiming to learn compact and generalizable 3D representation across scenes. By learning from in-the-wild 2D image collections with a volumetric renderer, our method takes significant steps toward generating scene-level 3D content without 3D annotations. We hope this may inspire future work in scene understanding and 3D-aware synthesis.

\paragraph{Limitations.} \textbf{1)} Our \scenerepresentation\ assumes the scene surface to be convex, which indicates that concave geometry, such as caves and tunnels, cannot be modeled and generated. \textbf{2)} Though our model does not require prior knowledge of the camera pose, further explorations are needed to increase the robustness of camera sampling during training. A naively random strategy sometimes leads to model collapse. \textbf{3)} Our model requires millions of training images and takes several days to converge. Future work on reducing the training cost would be fruitful.